\documentclass[]{fairmeta_gr2}

\usepackage{colortbl}
\usepackage{mdframed}
\usepackage{fvextra}
\usepackage{algorithm}
\usepackage{algorithmic}
\usepackage{amsthm}
\usepackage{amssymb}
\usepackage{amsmath}
\usepackage{mathtools}
\usepackage{xspace}
\usepackage{textcomp}
\usepackage{fancyvrb}
\usepackage{wrapfig}
\usepackage{fontawesome5}

\microtypesetup{expansion=false}

\definecolor{doderblue}{RGB}{30,144,255}
\definecolor{lightblue}{RGB}{210, 220, 250}
\definecolor{tagblue}{RGB}{0, 0, 200}
\newcommand{\HiLight}{\textsc{HiLight}\xspace}
\newcommand{\hl}[1]{\textcolor{purple}{\textbf{#1}}}
\newcommand{\hlb}[1]{\textcolor{doderblue}{#1}}
\newcommand{\blue}{\cellcolor{lightblue}}
\newcommand{\green}{\cellcolor{green!10}}

\newcommand{\revcolor}{black}
\newcommand{\rev}[1]{{\color{\revcolor}#1}}
\newenvironment{revblock}{\color{\revcolor}}{}

\title{Learning Evidence Highlighting for Frozen LLMs}

\author[1, *]{Shaoang Li}
\author[1, *]{Yanhang Shi}
\author[2]{Yufei Li}
\author[2]{Mingfu Liang}
\author[2]{Xiaohan Wei}
\author[2]{Yunchen Pu}
\author[2]{Fei Tian}
\author[2]{Chonglin Sun}
\author[2]{Frank Shyu}
\author[2]{Sandeep Pandey}
\author[2]{Luke Simon}
\author[2, \dagger]{Xi Liu}
\author[1, \dagger]{Jian Li}

\affiliation[1]{Stony Brook University}
\affiliation[2]{Meta AI}

\contribution[*]{Joint First Author}
\contribution[\dagger]{Joint Correspondence Author}

\abstract{Large Language Models (LLMs) can reason well over focused inputs, yet often miss decisive evidence buried in long, noisy contexts. We introduce \HiLight, an Evidence Emphasis framework that turns evidence selection into a learned, non-destructive input-side control problem for frozen LLMs. Rather than retrieving, pruning, compressing, or rewriting the input, \HiLight trains a lightweight Emphasis Actor to insert minimal highlight tags around pivotal spans while preserving the original context. A frozen Solver then reasons over the emphasized input. The Actor is trained only from the Solver's downstream task reward, requiring no evidence labels, Solver gradients, logits, or internal activations. This yields a solver-compatible alternative to retrieval-style hard selection, context compression, and instance-level prompt rewriting. Across sequential recommendation and QA, \HiLight consistently improves over manual prompting and strong automated prompt-optimization baselines, with gains up to +27.5\% over manual instruction (MI) and +10.8\% over the strongest baseline on Amazon-Beauty. The learned emphasis policy transfers zero-shot to smaller and larger unseen Solvers across model families, including API-based Solvers, and its highlights align with human supporting evidence up to 0.78 F1. These results suggest that evidence selection can be learned as a reusable input-side control mechanism for frozen LLMs.}

\DeclareRobustCommand{\githubicon}{\faGithub\hspace{0.45em}}
\metadata[Code]{\githubicon\url{https://github.com/yh-sh/HiLight}}
\date{April 27, 2026}

\begin{document}

\maketitle

\section{Introduction}
\label{sec:introduction}

Large Language Models (LLMs) are increasingly deployed in settings that require reasoning over long and imperfect contexts, such as multi-document question answering~\cite{asai2024selfrag}, agentic workflows with accumulated traces~\cite{yao2023react,shinn2023reflexion}, and personalization over noisy user histories~\cite{conf/recsys/KusanoAT25,jiang2025recgpt}. Despite progress in scaling model capacity and context length, effective context utilization remains a bottleneck: LLMs can overlook decisive evidence when it is diluted among irrelevant, redundant, or outdated information~\cite{yang2025llm}. Empirical studies have documented this failure mode, including the ``Lost in the Middle'' effect, where models attend less reliably to information that is neither near the beginning nor the end of a long input~\cite{liu2024lost}. As a result, simply providing more context can hurt as much as it helps, especially when the context contains many plausible distractors. For example, in sequential recommendation a single preference cue may appear once in a long history, and in multi-hop QA a small bridging fact can be easy to miss even when both supporting passages are present.

A useful perspective is that long-context failures often arise from coupling two subproblems into one forward pass: \textit{evidence selection}, which identifies task-relevant spans amid distractors, and \textit{reasoning}, which performs inference over that evidence. Modern LLMs can be strong reasoners given clean, focused inputs, yet may still fail to isolate sparse evidence in low signal-to-noise regimes. This motivates an explicit decoupling: rather than forcing a single Solver to both sift through noise and reason, we introduce a lightweight module whose sole role is to \textit{surface evidence}, allowing a frozen Solver to focus on reasoning.
If this decoupling is successful, the resulting evidence cues should not only help a given Solver, but may also remain useful across \textit{different Solver families and scales}, since they reflect salient structure in the input rather than the idiosyncrasies of a single backbone.

\begin{figure*}[t]
\centering
\includegraphics[width=1.0\textwidth]{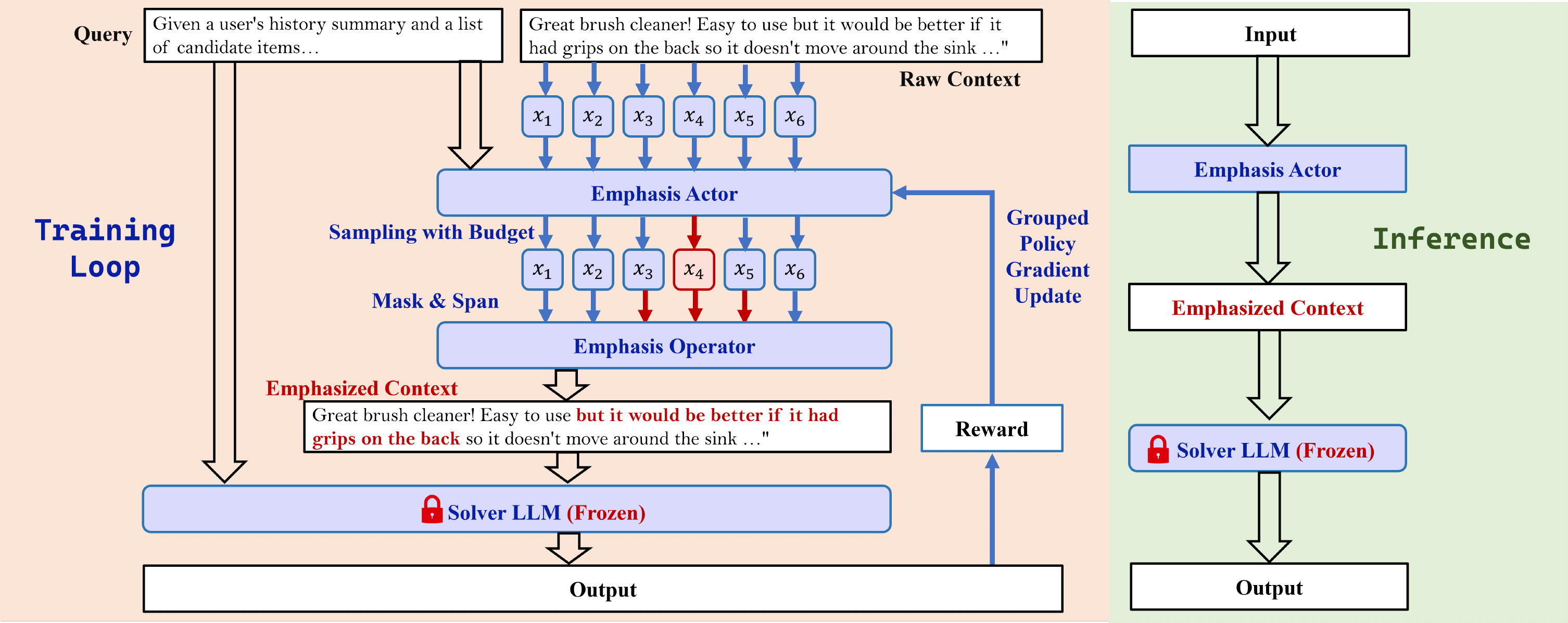}
\caption{\textbf{Overview of \HiLight.}
\HiLight decouples \textit{evidence selection} from \textit{reasoning} in long, noisy contexts.
\textbf{Inference:} Given query $Q$ and context $X$, a lightweight \textit{Emphasis Actor} selects pivotal spans under budget $\gamma$ and inserts minimal tags to form $\hat{X}$. A frozen \textit{Solver LLM} then produces the output.
\textbf{Training:} With no evidence annotations, we optimize the Actor using weakly supervised RL from the Solver reward $R(y,y^*)$, without Solver gradients or activations.
}
\label{fig:pipeline}
\end{figure*}

To this end, we propose \HiLight, an \textit{Evidence Emphasis} framework that learns \emph{instance-wise input markup} for long, noisy contexts. Inspired by how humans mark pivotal fragments when reading dense documents, \HiLight introduces a lightweight \textit{Emphasis Actor} that inserts minimal structural markers around selected spans in the \emph{original} context. The emphasized context is then consumed by a frozen \textit{Solver LLM} to produce the final answer or prediction (Figure~\ref{fig:pipeline}). Crucially, \HiLight does not compress, summarize, or rewrite the input: it preserves the raw text and intervenes only through lightweight, human-readable highlighting.

Thus, \HiLight studies a distinct design point: learning how to present the same fixed context to a frozen Solver, rather than changing which documents are retrieved, rewriting the context, or modifying the Solver. This matters because many context-management strategies trade fidelity for focus: retrieval and re-ranking select only a subset of passages~\cite{yu2024rankrag,zhao-etal-2024-longrag,sun2025dynamicrag}, pruning discards non-selected text~\cite{huang2025beyond}, compression and summarization may omit subtle evidence~\cite{jiang-etal-2023-llmlingua,jiang-etal-2024-longllmlingua,xu2024recomp,ge2024incontext,conf/nips/Mu0G23,fei2025efficient,wingate2022prompt}, and rewriting can alter the original input~\cite{weston2023system}. \HiLight instead treats evidence selection as learned, non-destructive input-side control: the Solver still receives the full context, while the Actor learns where to emphasize using only task reward. The central question is therefore not whether less context is better, but whether learned emphasis can improve how a frozen Solver uses the same context.

This design yields three practical benefits. First, it is \emph{solver-compatible}: \HiLight treats the Solver as frozen and requires only Solver outputs to compute reward, without gradients or activations, making it compatible with open-weight, API-based, or proprietary LLMs. Second, it is \emph{auditable}: the Actor produces explicit highlighted spans that expose what evidence was surfaced. Third, it is \emph{fidelity-preserving}: the original context remains intact, so surrounding information can still support disambiguation and multi-hop reasoning. As we later show, this input-side formulation also enables the learned emphasis policy to \textit{transfer zero-shot across Solver families and scales.}

While the high-level intuition is simple, it raises two tightly coupled questions:
\vspace{-0.09in}
\begin{tcolorbox}[colback=white!5!white,colframe=white!75!white]
\emph{How can we learn budgeted, instance-wise evidence highlighting for a frozen LLM using only downstream task feedback? When does non-destructive emphasis help more than hard selection or ordinary prompting?}
\end{tcolorbox}
\vspace{-0.1in}
We answer the first question with a weakly supervised learning formulation. The key operational challenge is that ``important evidence'' is typically latent and task dependent: most benchmarks provide only an input and an outcome label, not ground truth token level evidence. We therefore optimize the Emphasis Actor via reinforcement learning (RL) using only the Solver's downstream task reward. This enables the Actor to learn an evidence selection policy aligned with the task and Solver while requiring neither evidence annotations nor access to Solver internals.

We answer the second question empirically by studying \HiLight across tasks with different evidence sparsity and distractor structure.
Table~\ref{tab:overall} shows that \HiLight improves consistently over 
manual prompting and strong automated prompt-optimization baselines across sequential recommendation, multi-hop QA, reading comprehension, and biomedical QA classification. 
The gains are largest in high-distractor, sparse-signal settings: on Amazon-Beauty, \HiLight improves over manual instruction (MI) by +27.5\% HR@10 and +27.3\% NDCG@10, and over the strongest baseline by +10.8\% NDCG@10. Gains are smaller but consistent when evidence is more concentrated and the remaining bottleneck is complex inference. 
Consistent with this regime view, pruning to Actor-selected spans helps in sparse recommendation but hurts HotpotQA, while \HiLight improves both by preserving connective context.
Additional analyses show that \HiLight is stable across budgets and marker formats, transfers zero-shot to smaller/larger unseen Solvers including API-based Solvers, and learns highlights that substantially overlap with human-annotated evidence despite using no token-level supervision.

Our main contributions are summarized as follows:

$\bullet$ \textbf{Non-destructive input-side control.} We introduce \HiLight, an Evidence Emphasis framework that treats evidence selection as learned input-side control for frozen LLMs: it surfaces pivotal spans with minimal highlight tags while preserving the full original context.

$\bullet$ \textbf{Reward-only Actor learning.} We formulate instance-wise highlighting as a solver-compatible, weakly supervised RL problem and train a lightweight Emphasis Actor using only downstream task reward, without evidence annotations or access to Solver internals.

$\bullet$ \textbf{Empirical gains and regime characterization.} Across sequential recommendation and long-context QA benchmarks, \HiLight consistently outperforms manual prompting and strong prompt-optimization baselines, with the largest gains in high-distractor sparse-evidence regimes. Ablations show that non-destructive emphasis can outperform hard pruning when connective context matters.

$\bullet$ \textbf{Transferable and meaningful evidence.} A single Actor transfers zero-shot to smaller and larger unseen Solvers across model families, including API-based Solvers, while its highlights align with human-annotated evidence up to 0.78 F1 despite no token-level supervision.

\section{Problem Formulation}
\label{sec:formulation}

We formalize \emph{evidence emphasis} for long-context reasoning with a \emph{frozen} Solver LLM.

\textbf{Task setup.} Let $(Q, X, y^*)\sim\mathcal{D}$ denote a data instance, where $Q$ is a query (or instruction),  $X=(x_1,x_2,\dots,x_L)$ is a tokenized context of length $L$, $y^*$ is the ground-truth output. %
A frozen Solver LLM $\mathcal{M}$ maps $(Q,X)$ to an output $y$ (deterministically or stochastically). Performance is measured by a task-specific utility $R(y,y^*)$ such as EM, F1, accuracy, HR@10, or NDCG@10.

\textbf{Latent evidence.} As contexts grow longer, task-relevant information is often diluted by irrelevant, redundant, or outdated content. 
Conceptually, we can view each instance as containing one or more \emph{latent} evidence subsets within the context.  Let $[L]=\{1,2,\ldots,L\}$ and define an (unobserved) \textit{evidence set}
\[
E\!:=\!\{i\!\in\![L]\!:\!\text{$x_i$ is part of task-relevant evidence for $(Q,\!X)$}\}.
\]
We emphasize that $E$ is \emph{not} observed during training, need not be unique, and may depend on both the task and the Solver; we assume no token-level evidence annotations.

\textbf{Evidence highlighting policy.} We learn an Actor policy $\pi_\theta(\cdot\mid Q,X)$ that selects which parts of the context to emphasize. For simplicity, we begin with token-level selection via a binary mask $M\in\{0,1\}^L$, where $M_i=1$ indicates token $x_i$ is emphasized.
Given $(X,M)$, we construct an emphasized context $\hat X = g(X,M),$ where $g$ is a deterministic, \emph{non-destructive} operator that inserts minimal highlight tags around selected tokens (or coalesced spans) while preserving the original token content and order.

\textbf{Highlight budget.} To prevent degenerate all-highlight solutions and to keep emphasis concise and auditable, we impose a budget:
\begin{equation}\label{eq:budget}
\sum_{i=1}^L M_i \le \gamma L,
\end{equation}
where $\gamma\in(0,1]$ controls the maximum fraction of emphasized tokens. In practice, we enforce this constraint via a projection $\mathrm{Proj}_k$ with $k=\lfloor \gamma L\rfloor$.

\textbf{Objective.} For each instance, the Actor first samples a preliminary mask $\tilde M\sim\pi_\theta(\cdot\mid Q,X)$, projects it to a feasible mask $M=\mathrm{Proj}_k(\tilde M)$, and feeds the emphasized context $g(X,M)$ to the frozen Solver. The learning objective is
\begin{equation}
\label{eq:objective}
\mathcal{J}(\theta)=\mathbb{E}\left[R\left(y,y^*\right)\right],\quad y \sim \mathcal{M}\!\left(\cdot \mid Q, g(X,M)\right),
\end{equation}
where the expectation is over $(Q,X,y^*)\sim\mathcal{D}$, $\tilde M\sim\pi_\theta(\cdot\mid Q,X)$, the projection $M=\mathrm{Proj}_k(\tilde M)$, and any stochasticity of $\mathcal{M}$. If the Solver is deterministic, the final source of randomness is absent. Since evidence masks are latent and $\mathcal{M}$ is treated as a black box, $\mathcal{J}(\theta)$ cannot be optimized by backpropagating through the Solver. We therefore optimize the Actor with policy-gradient methods using only the downstream reward $R(y,y^*)$.

\section{\HiLight Method}
\label{sec:methodology}

This section instantiates the formulation in Section~\ref{sec:formulation}. Given a frozen Solver LLM $\mathcal{M}$ and a non-destructive emphasis operator $g(X,M)$, we describe (i) how \HiLight parameterizes the Actor policy $\pi_\theta(M\mid Q,X)$, (ii) how we enforce the highlight budget and construct emphasized inputs via span coalescence and tag injection, and (iii) how we optimize the Actor using task-level rewards.

\subsection{Emphasis Actor}
\label{sec:actor}

\textbf{Backbone and inputs.} The Actor $\pi_\theta$ is a pretrained LM backbone that encodes the concatenated input $[Q;X]$. Let $\mathbf{H}=(\mathbf{h}_1,\dots,\mathbf{h}_{L})$ denote the hidden states for the $L$ context tokens conditioned on $Q$\footnote{We implement the Actor with full attention over $(Q,X)$ to support evidence selection in long contexts, while the Solver remains frozen and can be open-weight or API-based.}.

\textbf{Token importance scores.} We map each context-token representation to an importance probability:
\begin{equation}\label{eq:importance_prob}
    p_i = \sigma\left(\frac{\mathbf{w}_p^\top \mathrm{LayerNorm}(\mathbf{h}_i)}{\tau}\right),
\end{equation}
where $\mathbf{w}_p\in\mathbb{R}^d$ is learnable, $\tau$ is a temperature, and $\sigma(\cdot)$ is the logistic function. This yields a probability map $P=\{p_1,\dots,p_L\}$ that defines a Bernoulli factorized policy over token selections.

\subsection{Budgeted Selection and Emphasis Construction}\label{sec:selection}

\textbf{Training-time sampling with budget projection.} The selected mask must satisfy the budget in Eq.~\eqref{eq:budget}. During training, we sample a preliminary mask $\tilde M\in\{0,1\}^L$ with $\tilde m_i\sim\mathrm{Bernoulli}(p_i)$, then project it to a feasible mask
\begin{equation}
\label{eq:projection}
M=\mathrm{Proj}_k(\tilde M),\qquad k=\lfloor \gamma L\rfloor,
\end{equation}
where $\mathrm{Proj}_k$ retains at most $k$ selected tokens, prioritizing larger $p_i$ values. This preserves exploration while guaranteeing budget feasibility; the projection is treated as part of the environment transition and is not differentiated through.

\textbf{Inference-time deterministic top-$k$.} At inference, we use a deterministic top-$k$ policy for stability, setting $M_i=1$ for the $k$ largest probabilities and $0$ otherwise.

\textbf{Span coalescence and Solver prompting.} Given $M$, the emphasis operator $g$ merges adjacent selected tokens into spans, optionally bridges short gaps of at most $\delta$ tokens, and inserts boundary tags such as \texttt{<start\_important>} and \texttt{<end\_important>} at span endpoints. The resulting emphasized context $\hat X=g(X,M)$ preserves the original text content and order. The frozen Solver receives $(Q,\hat X)$ under a fixed prompt template that tells it to treat tagged spans as emphasized evidence while still allowing it to use the surrounding context. Since \HiLight only changes the input string and never accesses Solver gradients, logits, or intermediate activations, it is compatible with black-box and API-based Solvers. Full prompt templates are provided in Appendix~\ref{app:prompts}.

\subsection{Actor Optimization with Task-level Reward}\label{sec:rl}

We freeze $\mathcal{M}$ and optimize only $\pi_\theta$. Because mask decisions are discrete and task metrics are generally non-differentiable, we train the Actor with a grouped policy-gradient objective using the downstream task metric as reward.

\textbf{Grouped policy gradient.} For each training instance, we sample $G$ preliminary masks $\{\tilde M_j\}_{j=1}^G$, project each to $M_j=\mathrm{Proj}_k(\tilde M_j)$, construct $\hat X_j=g(X,M_j)$, query the Solver, and obtain rewards
\begin{align}
r_j = R(y_j,y^*), \qquad y_j \sim \mathcal{M}(\cdot\mid Q,\hat X_j).
\end{align}
We normalize rewards within the group to form advantages
\begin{equation}\label{eq:advantage}
\hat A_j = \frac{r_j-\mu_r}{\sigma_r+\epsilon},
\end{equation}
where $\mu_r$ and $\sigma_r$ are the mean and standard deviation of $\{r_j\}_{j=1}^G$. The policy-gradient loss is
\begin{equation}\label{eq:pg_loss}
\mathcal{L}_{\mathrm{PG}}= -\frac{1}{G}\sum_{j=1}^{G}\hat A_j\log \pi_\theta(\tilde M_j\mid Q,X),
\end{equation}
where the log-probability is computed under the pre-projection Bernoulli policy; the projected masks are used only to construct emphasized inputs and evaluate rewards.

\textbf{Regularization.} To stabilize training, we add two regularizers. First, an entropy bonus encourages exploration and avoids premature collapse:
\begin{equation}\label{eq:ent_bonus}
\mathcal{L}_{\mathrm{ENT}}=-\mathcal{H}\!\left(\pi_\theta(\cdot\mid Q,X)\right),
\end{equation}
where $\mathcal{H}$ is the token-wise Bernoulli entropy averaged over policy-controlled tokens. Second, a soft length regularizer encourages the expected highlight fraction to match the budget:
\begin{equation}\label{eq:len_reg}
\mathcal{L}_{\mathrm{LEN}}=\left(\frac{1}{|\Omega|}\sum_{i\in\Omega} p_i-\gamma\right)^2,
\end{equation}
where $\Omega$ denotes valid policy-controlled context tokens. The entropy term maintains exploration, while the length term stabilizes the expected selection density before projection.

\textbf{Final loss.} The final objective minimized by the Actor is
\begin{equation}\label{eq:final_loss}
\mathcal{L}(\theta)=\mathcal{L}_{\mathrm{PG}}+\lambda_{\mathrm{len}}\mathcal{L}_{\mathrm{LEN}}+\beta_{\mathrm{ent}}\mathcal{L}_{\mathrm{ENT}}.
\end{equation}
We optimize $\theta$ with standard first-order optimizers. Appendix~\ref{app:loss_ablation} ablates the entropy and length-regularization terms, showing that removing either term degrades performance.
Algorithmic details and the exact Solver prompt template are provided in Appendix~\ref{app:prompts}.

\section{Experiments}
\label{sec:experiments}

We evaluate \HiLight on long-context tasks spanning different evidence sparsity and distractor regimes. Our experiments ask four questions: (i) Does instance-wise evidence emphasis improve frozen-Solver performance over strong prompt-optimization baselines?
(ii) Is \emph{non-destructive} emphasis preferable to hard pruning? (iii) What is the per-query inference and training cost? and (iv) Does the learned emphasis policy transfer across Solver families and produce meaningful evidence highlights?

\subsection{Experimental Setup}

\textbf{Tasks and datasets.}
We report results on multiple tasks spanning different sparsity regimes. Our evaluation includes Amazon-Beauty~\cite{hou2024bridging}, a widely used sequential recommendation dataset containing user review histories. For Amazon-Beauty, each example consists of a user history, one held-out ground-truth next item, and a fixed candidate set constructed following the standard evaluation protocol of~\cite{hou2024bridging}; HR@10 and NDCG@10 are computed over this candidate set. Details on candidate construction, negative sampling, and prompt formatting are provided in Appendix~\ref{app:benchmarks}. We also evaluate on HotpotQA~\cite{yang2018hotpotqa}, a dataset designed for multi-hop question answering; SQuAD 2.0~\cite{conf/acl/RajpurkarJL18}, a reading comprehension benchmark that includes unanswerable questions; and the artificial subset of PubMedQA~\cite{jin2019pubmedqa}, a biomedical dataset where the task is to answer research questions (Yes/No) based on abstracts.

\textbf{Solvers, Actors, and rewards.}
Unless otherwise specified, we use Qwen3-14B~\cite{yang2025qwen3} as the frozen Solver. The Actor is initialized from Qwen3 models: 0.6B for SQuAD 2.0/PubMedQA, 1.7B for Amazon-Beauty, and 4B for HotpotQA.  Appendix~\ref{app:sensitivity} provides a controlled Actor-size study, showing that larger Actors help most on structurally complex evidence-selection tasks such as HotpotQA, with diminishing returns beyond mid-sized Actors.
The highlight budget is $\gamma=0.25$ for Amazon-Beauty and $\gamma=0.15$ for the QA tasks. We optimize with Adam using learning rate $10^{-4}$, weight decay $10^{-2}$, group size $G=4$, entropy coefficient $\beta_{\mathrm{ent}}=1.0$, and length-regularization weight $\lambda_{\mathrm{len}}=0.01$. Training rewards are task-specific: a weighted HR@10/NDCG@10 reward for Amazon-Beauty, F1/EM for HotpotQA and SQuAD 2.0, and accuracy for PubMedQA; full details are in Appendix~\ref{app:exp_details}. \rev{Each Actor is trained in one pass over 3,000 (Amazon-Beauty) or 10,000 (QA) training examples, with the checkpoint selected on a disjoint 500-example training subset (Appendix~\ref{app:protocol}).}

\textbf{Baselines.}
We compare against MI and strong black-box prompt-optimization baselines: PRL~\cite{batorski2025prl}, BFRS~\cite{soylu2024fine}, OPRO~\cite{conf/iclr/Yang0LLLZC24}, DSPy/MIPROv2~\cite{conf/emnlp/Opsahl-OngRPBPZ24}, and APE~\cite{zhou2023large}. All methods use the same frozen Solver and the same raw context. Full prompt templates and configurations are in Appendix~\ref{app:configs}.

\subsection{Main Results}
\label{subsec:overall}

Table~\ref{tab:overall} shows that \HiLight achieves the best reported performance across all four tasks and eight metrics, outperforming both manual instruction (MI) and strong prompt-optimization baselines. This supports our central hypothesis: \emph{instance-wise emphasis on the data context} is a complementary and often stronger control surface than optimizing a single system prompt shared across inputs.

\begin{table*}[t]
\centering
\small
\setlength{\tabcolsep}{8pt}
\resizebox{\textwidth}{!}{
\begin{tabular}{lcccccccc}
\toprule
\multirow{2}{*}{\textbf{Baseline}} & \multicolumn{2}{c}{\textbf{Amazon-Beauty}}  & \multicolumn{2}{c}{\textbf{HotpotQA}}  & \multicolumn{2}{c}{\textbf{SQuAD 2.0}} & \multicolumn{2}{c}{\textbf{PubMedQA}}   \\
\cmidrule(lr){2-3}\cmidrule(lr){4-5}\cmidrule(lr){6-7}\cmidrule(lr){8-9}
& HR@10 & NDCG@10 & EM & F1 & EM & F1 & Acc. & M-F1 \\
\midrule
MI              & 0.02256 & 0.01247 &  0.556 & 0.706 & 0.586 & 0.647 & 0.921 & \hlb{0.776} \\
PRL                                  & 0.02692 & \hlb{0.01432} & 0.571 & 0.719 & \hlb{0.644} & 0.673 & 0.924 & 0.735 \\
BFRS                                 & 0.02363 & 0.01248 & \hlb{0.593} & 0.700 & 0.573 & \hlb{0.694} & 0.922 & 0.761\\
OPRO                                 & 0.02332 & 0.01346 & 0.591 & \hlb{0.723} & 0.620 & 0.672 & 0.922 & 0.768\\
DSPy (MIPROv2)                       & \hlb{0.02730} & 0.01410 & 0.565 & 0.721 & 0.590 & 0.685 & \hlb{0.930} & 0.767\\
APE                                  & 0.02390 & 0.01311 & 0.581 & 0.722 & 0.592 & 0.669 & 0.926 & 0.727\\
\midrule
\HiLight         &  \hl{0.02877} & \hl{0.01587} & \hl{0.606} & \hl{0.741} & \hl{0.661} & \hl{0.721} & \hl{0.940} & \hl{0.821}  \\
$\Delta_{\%}$(vs.\ Best Base.) & \green{+5.38\%} & \green{+10.82\%} & \green{+2.19\%} & \green{+2.49\%} & \green{+2.64\%} & \green{+3.89\%} & \green{+1.08\%} & \green{+5.80\%}  \\
$\Delta_{\%}$(vs.\ MI) & \blue{+27.53\%} & \blue{+27.27\%} & \blue{+8.99\%} & \blue{+4.96\%} & \blue{+12.80\%} & \blue{+11.44\%} & \blue{+2.06\%} & \blue{+5.80\%}  \\
\bottomrule
\end{tabular}
}
\caption{Overall performance comparison. Best results are highlighted in \hl{purple} and second-best results are marked in \hlb{blue}. %
\rev{$\Delta_{\%}$ is the relative improvement $100\,(m_{\HiLight}-m_b)/m_b$ over baseline $b$.}
\HiLight consistently outperforms all baselines across sequential recommendation, multi-hop QA, reading comprehension, and biomedical QA classification tasks. }
\label{tab:overall}
\end{table*}

\textbf{Where the gains are largest: sparse evidence under heavy distractors.}
The largest gains appear on Amazon-Beauty, where user histories are long and noisy and the decisive signal is sparse. \HiLight improves over the strongest baseline by +5.38\% HR@10 and +10.82\% NDCG@10, and over MI by +27.53\% HR@10 and +27.27\% NDCG@10. This is precisely the regime where evidence selection is the dominant bottleneck, consistent with \HiLight's goal of surfacing pivotal spans without rewriting the context. On HotpotQA, SQuAD 2.0, and PubMedQA, gains are smaller but consistent (1.08\%--5.80\% over the best baseline), suggesting that explicit emphasis can still improve evidence utilization when reasoning remains a substantial part of the task.

\textbf{Quality of evidence utilization.} 
Across tasks, \HiLight often yields relatively larger improvements on ranking- and overlap-sensitive metrics (NDCG@10, F1, M-F1) than on coarse binary metrics (HR@10, EM, Acc.). This is consistent with emphasis improving \emph{evidence utilization quality}: in recommendation, surfaced signals can move the ground-truth item upward within the top-$K$ list (reflected by NDCG@10) even when it was already a ``hit''; in QA, emphasizing supporting spans can produce more complete answer spans and higher token overlap (reflected by F1/M-F1) even when exact match is unchanged. Thus, \textit{\HiLight tends to improve how the Solver uses relevant evidence, not only whether a subset of outputs flip from incorrect to correct.}

\subsection{Why Non-Destructive Emphasis?}
\label{subsec:ablation}

To isolate the role of non-destructive emphasis, Table~\ref{tab:ablation} compares \HiLight with four controls: MI (no highlighting), Random highlighting under the same budget, Few-shot highlighting demonstrations, and Pruned, which feeds only Actor-selected spans to the Solver while discarding the rest.

\textbf{Sparse-evidence recommendation.} 
On Amazon-Beauty, %
Random highlighting slightly improves over MI, and few-shot exemplars improve further, consistent with the idea that \textit{long noisy histories benefit from explicit focus cues that help the Solver focus on a subset of relevant signals.}  The Pruned variant is strong (0.02740 HR@10, 0.01448 NDCG@10), showing that restricting the Solver to the Actor-selected spans can already recover much of the predictive signal in this sparse-evidence regime. However, \HiLight performs best, indicating that \emph{non-destructive emphasis} is advantageous: \textit{highlighting steers the Solver toward key spans while preserving surrounding context that can help with disambiguation and preference consistency.}

\begin{wraptable}{r}{0.6\textwidth}
\centering
\small
\setlength{\tabcolsep}{8pt}
\begin{tabular}{lcccc}
\toprule
\multirow{2}{*}{\textbf{Variant}} 
& \multicolumn{2}{c}{\textbf{Amazon-Beauty}} 
& \multicolumn{2}{c}{\textbf{HotpotQA}} \\
\cmidrule(lr){2-3}\cmidrule(lr){4-5}
& \textbf{HR@10} & \textbf{NDCG@10} & \textbf{EM} & \textbf{F1} \\
\midrule
MI       & 0.02256 & 0.01247 & \hlb{0.556} & \hlb{0.706} \\
Random        & 0.02380 & 0.01270 & 0.521 & 0.635 \\
Few-shot& 0.02403 & 0.01315 & \hlb{0.556} & 0.695 \\
Pruned         & \hlb{0.02740} & \hlb{0.01448} & 0.525 & 0.658 \\
\midrule
\HiLight                & \hl{0.02877} & \hl{0.01587} & \hl{0.606} & \hl{0.741} \\
\bottomrule
\end{tabular}
\caption{Ablation study on Amazon-Beauty and HotpotQA tasks. Best results are highlighted in \hl{purple} and second-best results are marked in \hlb{blue}. %
}
\label{tab:ablation}
\end{wraptable}

\textbf{Context preservation for reasoning.} 
A contrasting pattern appears on HotpotQA: Random highlighting, Few-shot highlighting, and especially Pruned context do not improve over MI, and pruning substantially hurts EM. This suggests that multi-hop QA is sensitive to preserving \emph{connective} material beyond the final supporting facts, such as bridging entities, discourse cues, and local context that help the Solver link evidence across passages. \HiLight is the only variant that improves EM and yields the strongest overall gains, supporting our design choice to \emph{emphasize without deleting}: focus cues help, but removing surrounding context can undermine multi-step reasoning even when the selected spans are informative.

\textbf{Relation to within-context retrieval.} This ablation also clarifies the relation to retrieval-style baselines in a fixed-context setting. Once the full context is already provided to the Solver, chunk retrieval or reranking within that context reduces to \emph{hard selection}: scoring chunks or spans, retaining the top-$k$, and discarding the rest. The Pruned ablation directly probes this family. Results show that hard selection can help when the signal is sparse and distractor-heavy, but can hurt reasoning-heavy QA when connective context is removed. This motivates \HiLight's non-destructive design. \rev{Hard selection remains below \HiLight on HotpotQA even when the selector is not the \HiLight Actor: an off-the-shelf reranker that keeps the top-$k$ sentences reaches 0.568/0.710 EM/F1, and a pruning selector trained with the same Actor architecture, RL procedure, reward, and Solver-query budget reaches 0.563/0.707, compared with 0.606/0.741 for \HiLight (Appendix~\ref{app:controls}).}

\subsection{Long-Context Stress Test}
\label{subsec:longctx}

\begin{table}[t]
\centering
\small
\begin{tabular}{lcccc}
\toprule
Context length & $\sim$2K & $\sim$8K & $\sim$16K & $\sim$32K \\
\midrule
$\Delta$ EM & +9.0\% & +5.0\% & +4.6\% & +3.9\% \\
$\Delta$ F1 & +5.0\% & +4.0\% & +2.7\% & +4.0\% \\
\bottomrule
\end{tabular}
\vspace{4pt}
\caption{HotpotQA stress test with injected distractors. Values are relative gains over MI; full results are in Appendix~\ref{app:sensitivity}.}
\label{tab:longctx_compact}

\vspace{4pt}
\setlength{\tabcolsep}{4pt}
\begin{tabular}{lcccccccc}
\toprule
Method & MI & PRL & BFRS & \begin{tabular}[c]{@{}c@{}}DSPy\\(MIPROv2)\end{tabular} & APE & OPRO & \begin{tabular}[c]{@{}c@{}}Few-shot with\\highlights\end{tabular} & \HiLight \\
\midrule
Input tokens & 1.0$\times$ & 1.0$\times$ & 3.0$\times$ & 1.0$\times$ & 1.0$\times$ & 1.0$\times$ & 3.0$\times$ & $<$ 1.01$\times$ \\
Training Solver queries & 0 & 120K & 60K & 24K & 60K & 30K & 0 & 12K \\
\bottomrule
\end{tabular}
\vspace{4pt}
\caption{Efficiency and Solver cost on Amazon-Beauty. Input tokens are normalized to MI; training cost counts Solver queries.}
\label{tab:efficiency}
\end{table}
To test whether \HiLight remains useful as distractor load increases, we extend HotpotQA inputs by injecting distractor documents to create contexts of approximately 8K, 16K, and 32K tokens. Table~\ref{tab:longctx_compact} reports relative gains over MI at each length; full absolute EM/F1 values are provided in Appendix~\ref{app:sensitivity}. \HiLight maintains positive gains throughout, supporting the view that the method targets evidence sparsity under distractors rather than only a particular benchmark length.

\subsection{Efficiency: Tokens, Solver Queries, and Latency}
\label{subsec:efficiency}

Table~\ref{tab:efficiency} reports efficiency on Amazon-Beauty. We measure \emph{Solver-side cost} in two complementary ways: (i) input-token overhead at inference, and (ii) the number of \emph{Solver queries} used during optimization, which is the dominant shared expense across black-box prompt-optimization methods.

\textbf{Minimal Solver-token overhead at inference.}
At inference, \HiLight uses one lightweight Actor pass followed by a single Solver call per instance. Because \HiLight only inserts short boundary markers around selected spans, the Solver input length increases by less than $1.01\times$ while preserving the full raw context. In contrast, in-context baselines that rely on demonstrations (e.g., BFRS and Few-shot) expand the Solver input to $3.0\times$, directly reducing the usable context window for long histories. Thus, \HiLight improves context utilization without paying the large input-length cost of demonstration-based prompting.

\textbf{Favorable accuracy--Solver-query trade-off.}
Search-based prompt optimizers spend many Solver calls evaluating candidate instructions, demonstrations, or prompt programs. \HiLight instead uses Solver calls to train a token-level evidence selector and reaches strong performance with 12K Solver queries, compared with 24K for DSPy, 30K for OPRO, 60K for APE/BFRS, and 120K for PRL. On Amazon-Beauty, this corresponds to a $\sim 5$--$10\times$ lower Solver-query budget than APE/PRL while also achieving higher accuracy, placing \HiLight on a favorable accuracy--Solver-cost frontier.

\textbf{Wall-clock latency is dominated by the Solver.}
We additionally measure end-to-end wall-clock latency. Across Actor sizes, p95 Actor latency ranges from 0.057\,s (0.6B) to 0.635\,s (8B), while the 14B Solver takes 9.1--19.6\,s depending on task and context length. The 4B Actor used in our most demanding setting adds under 0.34\,s at p95, corresponding to roughly 1.3\%--2.9\% of total pipeline latency. Full latency breakdowns are provided in Appendix~\ref{app:latency}.

\textbf{Scope of the efficiency claim.}
Our token and latency measurements are for \emph{per-query} fixed-context inference, where each query induces one Actor pass and one Solver call. Conversational multi-turn serving with cache-aware reuse across changing query-conditioned highlights would require an additional systems layer and is left to future work.

\subsection{Transferability across Solver Families and Scales}
\label{subsec:transfer}

A practical and conceptually important question is whether the evidence-selection policy learned by the Actor is tied to the particular Solver used during training, or whether it remains useful when deployed with substantially different Solvers. This question directly tests the premise behind \HiLight: if evidence selection can be partly decoupled from reasoning, then the learned emphasis policy should capture reusable structure in the input rather than only the idiosyncrasies of one Solver.

\textbf{Zero-shot transfer setup.}
We train the Emphasis Actor once with a \emph{source} Solver, Qwen3-14B, and apply it \emph{zero-shot} to five unseen \emph{target} Solvers spanning different families and scales: Qwen3-4B, Qwen3-8B, Qwen3-32B, Gemma-3-27B, and Llama-3-70B. Neither the Actor nor the target Solver is updated. For each target Solver, we compare against that Solver's own MI baseline, so the reported gains reflect transfer relative to the target model's native performance.

\textbf{Transfer across both smaller and larger Solvers.}
Table~\ref{tab:transfer} shows consistent gains across all five target Solvers, including both smaller and larger models than the training Solver. \HiLight improves HR@10 by +8.78\% to +19.17\% and NDCG@10 by +8.48\% to +31.56\% relative to each target Solver's own MI baseline. The gains on Llama-3-70B and Gemma-3-27B are especially notable: even substantially larger Solvers, which should already be stronger at filtering distractors, still benefit from explicit evidence emphasis. This suggests that the bottleneck is not only raw reasoning capacity, but also reliable localization and use of sparse evidence in noisy context.

\begin{table}[t]
\centering
\small
\setlength{\tabcolsep}{4pt}
\begin{tabular}{llccccc}
\toprule
\multirow{2}{*}{\textbf{Metric}} 
& \multirow{2}{*}{\textbf{Method}} 
& \multicolumn{2}{c}{\textbf{Smaller than Source}} 
& \multicolumn{3}{c}{\textbf{Larger than Source}} \\
\cmidrule(lr){3-4}\cmidrule(lr){5-7}
& & Qwen3-4B & Qwen3-8B & Gemma-3-27B & Qwen3-32B & Llama-3-70B \\
\midrule
\multirow{2}{*}{\textit{HR@10}}
& Self-Mark & +3.12\% & +3.81\% & +7.86\% & +5.13\% & +6.50\% \\
& \HiLight   & +13.06\% & +13.67\% & +15.77\% & +8.78\% & +19.17\% \\
\midrule
\multirow{2}{*}{\textit{NDCG@10}}
& Self-Mark & +5.23\% & +0.45\% & +7.48\% & +6.97\% & +9.12\% \\
& \HiLight   & +8.48\% & +12.43\% & +31.56\% & +21.24\% & +27.37\% \\
\bottomrule
\end{tabular}
\vspace{0.1in}
\caption{Cross-Solver transfer on Amazon-Beauty ($\Delta_{\%}$ over each target Solver's own MI baseline). Actor trained with Qwen3-14B and applied zero-shot.}
\vspace{-0.2in}
\label{tab:transfer}
\end{table}

\textbf{Comparison with self-generated highlighting.}
We further compare against \textsc{Self-Mark}, where the \emph{target} Solver is prompted to first highlight evidence in the context and then answer using its own self-generated markup. Although \textsc{Self-Mark} gives the target Solver an additional self-highlighting step, \HiLight outperforms it across all target Solvers and both metrics. This indicates that a specialized evidence selector trained from task-level feedback can produce higher-quality and more consistent emphasis cues than zero-shot self-highlighting, even when the target Solver is much larger than the Actor-training Solver. In particular, \textsc{Self-Mark} yields weaker and less stable NDCG@10 gains (e.g., only +0.45\% on Qwen3-8B), whereas \HiLight consistently improves ranking quality. We attribute this gap to the fact that self-highlighting relies on the target Solver's generic instruction-following, which may not align with downstream task reward, %
while \HiLight is trained to select spans that are actually useful for that reward. %
The larger and more stable NDCG@10 gains further echo Table~\ref{tab:overall}: learned emphasis improves ranking quality, not merely binary hit-or-miss retrieval.

\textbf{Tokenizer-independent transfer via text spans.}
Cross-Solver transfer does not require vocabulary or token alignment between the Actor and the target Solver. The Actor's token-level scores are projected back to character-level spans using tokenizer offset mappings. Boundary markers are then inserted into the raw text string, and the target Solver tokenizes this modified text using its own tokenizer. Thus, the transferred object is the emphasized \emph{text span}, not the Actor's token IDs. This also explains why \HiLight is compatible with Solvers from different model families and why marker-format robustness in Appendix~\ref{app:sensitivity} supports the claim that gains are not tied to a brittle delimiter tokenization.

\begin{table}[h]
\centering
\small
\setlength{\tabcolsep}{4pt}
{\color{\revcolor}
\begin{tabular}{lcccccc}
\toprule
\textbf{Metric}
& \begin{tabular}[c]{@{}c@{}}GPT-5\\mini\end{tabular}
& \begin{tabular}[c]{@{}c@{}}GPT-5\\(low)\end{tabular}
& \begin{tabular}[c]{@{}c@{}}GPT-5\\(high)\end{tabular}
& \begin{tabular}[c]{@{}c@{}}Claude\\Haiku 4.5\end{tabular}
& \begin{tabular}[c]{@{}c@{}}Claude\\Sonnet 5\end{tabular}
& \begin{tabular}[c]{@{}c@{}}Claude\\Opus 4.8\end{tabular} \\
\midrule
\textit{EM} & +1.70\% & +0.59\% & +0.29\% & +17.78\% & +2.17\% & +1.47\% \\
\textit{F1} & +1.11\% & +1.00\% & +0.98\% & +19.39\% & +1.76\% & +1.57\% \\
\bottomrule
\end{tabular}}
\vspace{0.1in}
\caption{\rev{Zero-shot transfer to API-based Solvers on HotpotQA ($\Delta_{\%}$ over each Solver's own MI baseline).}}
\label{tab:api_solvers}
\end{table}

\rev{\textbf{Transfer to API-based Solvers and across tasks.}
Without adaptation, the same Actor improves EM and F1 for all six API-based Solvers in Table~\ref{tab:api_solvers}, with smaller gains for stronger Solvers within each family (Appendix~\ref{app:api_solver}). The HotpotQA Actor applied unchanged to SQuAD 2.0 also improves EM by +3.58\% and F1 by +4.64\% over MI, compared with +12.80\% and +11.44\% for the task-specific Actor (Appendix~\ref{app:cross_task}).}

\textbf{Takeaway.}
Overall, these transfer results support a key premise of \HiLight: a substantial portion of evidence-selection knowledge reflects task and input structure rather than only the idiosyncrasies of a single Solver. The learned emphasis policy transfers to smaller Solvers, larger Solvers, different model families\rev{, API-based Solvers, and a related task}, and %
Appendix~\ref{app:sensitivity} provides robustness checks across highlight budgets, delimiter formats, span-bridging thresholds, \rev{loss coefficients, and Solver prompt templates}.
Together, these results empirically support the view that evidence selection and reasoning are at least partially separable.

\subsection{Evidence Quality and Interpretability}
\label{subsec:evidence_quality}

A natural question is whether the learned highlights correspond to semantically meaningful evidence or merely exploit Solver-specific formatting biases. HotpotQA provides human-annotated supporting facts, allowing both qualitative and quantitative evaluation of evidence quality.

\begin{figure*}[t]
\centering

\begin{minipage}[t]{0.55\textwidth}
\vspace{0pt}
\centering
\includegraphics[width=\linewidth]{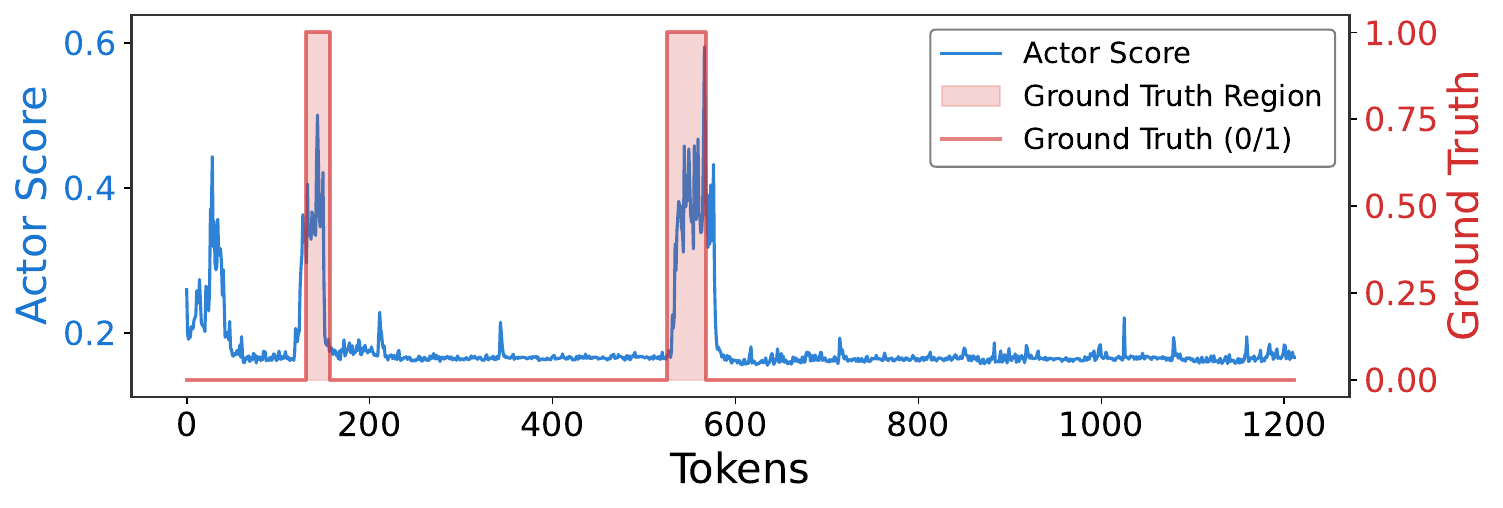}
\vspace{-0.25in}
\captionof{figure}{Token-level Actor importance scores on a HotpotQA example. Higher scores concentrate on a small set of regions, with substantial overlap with annotated evidence spans.}
\label{fig:viz}
\end{minipage}
\hfill
\begin{minipage}[t]{0.40\textwidth}
\vspace{0pt}
\centering
\small
\setlength{\tabcolsep}{4pt}
\begin{tabular}{lccc}
\toprule
\textbf{Actor Size} & \textbf{Precision} & \textbf{Recall} & \textbf{F1} \\
\midrule
0.6B & 0.7275 & 0.6433 & 0.6828 \\
1.7B & 0.7964 & 0.7016 & 0.7460 \\
4B   & 0.8367 & 0.7215 & 0.7748 \\
8B   & 0.8403 & 0.7307 & 0.7817 \\
\bottomrule
\end{tabular}
\captionof{table}{Token-level overlap between Actor highlights and ground-truth supporting facts on HotpotQA. The Actor is trained without evidence annotations.}
\label{tab:evidence_overlap}
\end{minipage}

\end{figure*}

\textbf{Actor scores concentrate on supporting evidence.}
Figure~\ref{fig:viz} visualizes token-level Actor scores on a representative HotpotQA instance. The scores form a small number of pronounced peaks, with high-scoring regions substantially overlapping the annotated supporting evidence. Between these peaks, scores remain low, indicating that the Actor uses its budget to focus on a limited set of candidate evidence spans rather than broadly reweighting the entire input.

\textbf{Quantitative evidence alignment.}
We threshold Actor scores at 0.5 to obtain predicted evidence tokens and compute precision, recall, and F1 against HotpotQA supporting-fact annotations. Table~\ref{tab:evidence_overlap} shows that the Actor achieves up to 0.78 F1 overlap with human evidence, despite never seeing evidence labels during training. The monotonic improvement from 0.68 F1 with a 0.6B Actor to 0.78 F1 with an 8B Actor mirrors the downstream scaling trend, suggesting that additional Actor capacity improves evidence identification rather than merely amplifying formatting effects.

Case studies and failure analysis are provided in Appendix~\ref{app:case}.

\section{Conclusion}
\label{sec:conclusion}

We introduced \HiLight, a learned evidence-emphasis framework that treats evidence selection as non-destructive input-side control for frozen LLMs. A lightweight Actor learns from task-level reward alone to insert minimal highlight markers into the original context, allowing a frozen Solver to reason over emphasized but unaltered input. Experiments across recommendation and QA show consistent gains over manual prompting and strong prompt-optimization baselines, especially in high-distractor regimes where evidence is sparse.  Transfer and evidence-alignment analyses further suggest that the Actor learns reusable, human-meaningful evidence structure rather than merely exploiting a single Solver's formatting preferences.
Our current study focuses on per-query fixed-context inference; cache-aware multi-turn serving, instance-adaptive highlight budgets, and broader rewriting-based comparisons remain important future directions. From a broader-impact perspective, \HiLight may improve the auditability and reliability of LLM-based decision pipelines by exposing which evidence was emphasized, but incorrect or biased emphasis could also mislead users in sensitive applications.

\rev{\textbf{Limitations.}
\HiLight assumes that the query and the context arrive as separate fields; the Actor conditions on both but highlights only the context, and in multi-turn use the highlights must be recomputed whenever either changes. Each task uses its own Actor trained from task reward, a per-task cost that cross-Solver and cross-task transfer reduce but do not remove, and all Actors in our study come from the Qwen3 family. Highlights can also fall on non-decisive spans, and correctly highlighted evidence does not guarantee a correct answer (Appendix~\ref{app:case}). Finally, our bootstrap intervals quantify aggregate sampling uncertainty rather than per-instance calibration, so a highlight is not a confidence estimate for an individual prediction.}

\section*{Acknowledgements}
The work of S. Li, Y. Shi and J. Li was supported in part by the National Science Foundation (NSF) grants 2148309, 2315614 and 2337914, and the National Institutes of Health (NIH) grant R01HL184139.  Any opinions, findings, and conclusions or recommendations expressed in this material are those of the authors and do not necessarily reflect the views of the funding agencies.

\bibliographystyle{assets/plainnat}
\bibliography{paper}

\clearpage
\beginappendix

\section{Related Work}

\textbf{Long-context utilization and evidence selection.}
A central challenge in long-context LLM reasoning is to ensure that task-relevant evidence is not diluted by distractors.
Most prior approaches manage long inputs by either reducing input volume or explicitly selecting salient regions.
A dominant line is \emph{hard selection}, such as retrieving top-$k$ passages, pruning history, or re-ranking candidate context segments before inference~\cite{yu2024rankrag,zhao-etal-2024-longrag,sun2025dynamicrag,huang2025beyond}. While effective, hard selection can discard surrounding text that is useful for disambiguation, negative reasoning, or maintaining discourse structure.  In our setting, the full context is already given to the Solver, so a retrieval-style method within a single context reduces to \emph{hard selection}: chunking the context, scoring chunks or spans, and retaining only the top-$k$ subset while discarding the rest. We study this family through our \emph{Pruned} ablation, which shows that deletion-based selection can help in sparse-signal recommendation but can hurt multi-hop QA when connective context is removed.

Another line reduces context length through prompt compression, learned gist tokens, summarization, or context distillation~\cite{jiang-etal-2023-llmlingua,jiang-etal-2024-longllmlingua,conf/nips/Mu0G23,xu2024recomp,ge2024incontext,fei2025efficient,wingate2022prompt}. These methods reduce length and can increase the density of relevant information, but they are inherently lossy and can introduce errors when the compressed representation omits subtle but decisive details. A related strategy prompts the LLM itself to extract or regenerate a distilled context before answering~\cite{weston2023system}, but this adds an extra generation step, still risks discarding bridging information, and couples evidence selection to the Solver's own instruction-following ability. Most relevant to our setting are methods that extract evidence or rationales from LLM internal signals, such as attention or attribution maps, to support interpretability or post-hoc explanation~\cite{yuksekgonul2024attention,conf/iclr/HalawiDS24,liu-etal-2025-selfelicit}. For example, SelfElicit~\cite{liu-etal-2025-selfelicit} highlights evidence at inference time using the Solver's own attention-derived signals, tying evidence selection to the target model's internals. \rev{Attention-steering methods share the goal of directing a model toward important input content: PASTA~\cite{zhang2024tell} reweights attention toward user-specified spans in selected heads, Selective Prompt Anchoring~\cite{tian2025selective} amplifies the influence of selected prompt tokens on the output logits, and SEKA~\cite{li2026spectral} edits key embeddings to steer attention toward highlighted text. These methods operate on attention or logits inside the model and take the emphasized spans as given, whereas \HiLight learns where to place emphasis from task reward.}
In contrast to approaches that require internal access or primarily analyze a fixed model after the fact, \HiLight performs an \emph{input-side intervention} that is compatible with frozen Solvers: it inserts explicit emphasis markers \emph{before} inference to steer the Solver toward pivotal spans while keeping the original context intact.

\textbf{Context markup and inference-time control.} A related and widely used practice for controlling LLM behavior is to introduce structured markup in the input, such as delimiters, section headers, XML wrappers, and other formatting conventions that shape how the model parses and attends to context~\cite{yao2023react,beurer2023prompting}. Such structured prompting can improve reliability by making evidence boundaries explicit and by separating instructions from data~\cite{yao2023react,schick2023toolformer}. Recent context-formatting work such as C-NORM further shows that context presentation choices, including delimiter and structural-marker conventions, can substantially affect long-context utilization~\cite{chen2025cnorm}. However, such methods typically optimize global or model-specific formatting choices, and they do not address the key challenge in long-context settings: deciding \emph{where} the decisive evidence is for each instance~\cite{li2023guiding}. \HiLight bridges this gap by learning \emph{instance-wise} placement of minimal highlight tags on top of an unaltered context, effectively providing a learned control surface for evidence emphasis. Orthogonally, model-internal approaches such as RetMask improve context utilization by modifying or post-training the Solver~\cite{ma2026retmask}, whereas \HiLight changes only the input string and remains compatible with frozen and API-based Solvers. Our robustness experiments further test whether gains persist across different tag formats and highlighting budgets, indicating that the method is not merely exploiting a brittle formatting quirk.

\textbf{Automatic Prompt Optimization (APO) and test-time scaling.}
Traditional prompt engineering relies on manual trial-and-error~\cite{liu2023pre}, while APO automates this process by treating prompts as learnable objects~\cite{zhang2023automatic}. Existing methods optimize instructions via gradient-based updates~\cite{yuksekgonul2025optimizing}, evolutionary search~\cite{pmlr-v235-fernando24a,wang2025evolving}, or RL~\cite{zhang2023tempera,mao2025reinforced}. A fundamental distinction is the optimization scope: APO typically searches for a single, static system prompt optimized at the \emph{task level}. In contrast, \HiLight performs \emph{instance-level} optimization on the \emph{data context} itself by dynamically deciding where to place saliency markers for each specific input, enabling adaptation to varying noise and evidence distributions across queries while consuming the same raw evidence. Orthogonally, test-time scaling strategies such as best-of-$N$ sampling or self-consistency can improve accuracy but often increase latency and compute; \HiLight instead aims to improve evidence utilization via a single emphasized input, without relying on extensive over-sampling.

\textbf{Token-level optimization.} 
The axiom that tokens possess unequal semantic weight is foundational to modern NLP, underpinning mechanisms from attention~\cite{corr/BahdanauCB14,conf/naacl/YangYDHSH16,conf/nips/VaswaniSPUJGKP17} to efficient inference~\cite{zhang2023h2o,xiao2024efficient}. 
Recent works exploit token non-uniformity to refine training objectives, reweight losses, or optimize token-level inputs~\cite{lin2024not,liu2025tisdpo,wu2025generalization,huang2025beyond}. Closest in spirit, reinforced token-level input refinement learns token-level edits or selections through RL, but its goal is input refinement rather than non-destructive evidence emphasis for a frozen Solver; refinement or pruning can change what information is passed forward. \HiLight instead externalizes token importance as explicit structural markers: it inserts emphasis cues into the raw input, preserves all original tokens, and trains a separate Actor from task-level reward without accessing or modifying the Solver.

\textbf{LLMs for recommendation and personalization.}
Recent work has explored using LLMs as recommender components, including zero-shot or few-shot recommendation via natural-language prompts~\cite{corr/abs-2303-14524,conf/recsys/KusanoAT25,conf/sigir/LiuY0ZGZLSG25,jiang2025recgpt},
modeling user histories and item identifiers in semantic space~\cite{conf/recsys/Geng0FGZ22,deng2025onerec,liu2025onerec,liang2026generative}, and hybrid systems that integrate LLM representations with classic recommenders and external tools~\cite{zhao2024let,zhang2025reasonrec}.
Despite promising results, LLM-based recommendation faces an extreme long-context regime: user histories can be ultra-long, noisy, and temporally non-stationary, so relevant signals are often sparse and easily overshadowed by distractors.
This makes recommendation a natural stress test for evidence emphasis, and our results show that \HiLight provides a lightweight and interpretable mechanism for improving long-context utilization in such noisy personalization settings.

{\color{red}

}

\section{Training Algorithm and Solver Prompt Template}\label{app:prompts}

\subsection{Frozen Solver Prompting}
\label{sec:solver}
\begin{figure}[h!]
    \centering
    \begin{tcolorbox}[
        colback=white,   
        colframe=black,  
        boxrule=0.5pt,  
        sharp corners,
        fontupper=\small
    ]
        \textbf{[CONTEXT]} \\
        \texttt{... the patient exhibited \textcolor{tagblue}{<start\_important>} symptoms of severe fatigue \textcolor{tagblue}{<end\_important>} ...} \\
        (The \texttt{\textcolor{tagblue}{<start\_important>}...\textcolor{tagblue}{<end\_important>}} tags indicate emphasized evidence.)
        
        \vspace{0.2cm}
        \hrule
        \vspace{0.2cm}
        
        \textbf{[INSTRUCTION]} \\
        \texttt{Answer the query using the context above.}
    \end{tcolorbox}
    \caption{The consistent prompt template used for the frozen Solver. The Emphasis Actor injects \textcolor{tagblue}{blue tags} into the raw context.}
    \label{fig:prompt_template_colored}
\end{figure}

The Solver LLM $\mathcal{M}$ receives $(Q,\hat{X})$ and generates an output $y$. We use a consistent prompt template that encourages attention to highlighted spans while still allowing the model to use surrounding context when needed. Because \HiLight only modifies inputs and never requires gradients or intermediate activations from $\mathcal{M}$, it is compatible with both open-weight and API-based Solvers.

\subsection{Actor Optimization with Task-level Reward}%

Algorithm~\ref{alg:training} summarizes the complete training procedure for \HiLight, including grouped sampling, budget projection, emphasis construction, and the Actor update under Eq.~\eqref{eq:final_loss}.

\begin{algorithm}[h!]
\caption{\HiLight training with grouped policy gradient 
(Solver frozen)}
\label{alg:training}
\begin{algorithmic}[1]
\STATE \textbf{Input:} Dataset $\mathcal{D}$, Actor $\pi_\theta$, frozen Solver $\mathcal{M}$, group size $G$, budget $\gamma$.
\WHILE{not converged}
    \STATE Sample a batch $B=\{(Q,X,y^*)\}\sim\mathcal{D}$.
    \FOR{each $(Q,X,y^*)$ in $B$}
        \FOR{$j=1$ to $G$}
            \STATE Compute $P=\{p_i\}$.
            \STATE Sample $\tilde{M}_j$ via Bernoulli$(p_i)$ and project to budget $k=\lfloor \gamma L \rfloor$ to obtain $M_j$.
            \STATE Construct $\hat{X}_j = g(X,M_j)$ (span coalescence + tag insertion).
            \STATE Query Solver $y_j \leftarrow \mathcal{M}(Q,\hat{X}_j)$ and compute reward $r_j \leftarrow R(y_j,y^*)$.
        \ENDFOR
        \STATE Compute $\{\hat{A}_j\}$ by normalizing $\{r_j\}$ within the group (Eq.~\eqref{eq:advantage}).
        \STATE Update Actor parameters $\theta$ by minimizing Eq.~\eqref{eq:final_loss}. 
    \ENDFOR
\ENDWHILE
\STATE \textbf{Output:} Optimized Actor $\pi_\theta$.
\end{algorithmic}
\end{algorithm}

\section{Benchmarks}
\label{app:benchmarks}

To evaluate the effectiveness of our framework, we compare against a broad suite of baselines spanning manual engineering and state-of-the-art automated optimization methods:
\begin{itemize}
    \item \textbf{MI} (Manual Instruction). A foundational baseline in which the Solver LLM is prompted with human-written instructions based on intuition, without automated optimization or any structural emphasis applied to the context.
    
\item \textbf{PRL} \cite{batorski2025prl}. A reinforcement learning--based framework that trains a prompt generator to iteratively improve instructions by producing an intermediate reasoning trace before emitting the final prompt, and autonomously synthesizes and integrates task-specific few-shot examples to boost the performance of a frozen evaluation model. \emph{LLM-call cost:} with 3{,}000 training datapoints, PRL typically requires (i) running the evaluation LLM to obtain rewards/scores and (ii) additional generator calls to sample candidate prompts (often with multiple rollouts per update). These repeated score--and--update cycles over the training set (and sometimes across multiple epochs/mini-batches) lead to the largest total number of calls.

\item \textbf{BFRS} \cite{soylu2024fine}. An optimization strategy for modular language-model pipelines that performs best-first search over prompt candidates and iteratively refines them using feedback from model executions. By bootstrapping training data from the model's own successful executions, BFRS enables self-improvement. \emph{LLM-call cost:} BFRS expands a search tree of candidates; each expansion requires scoring many nearby prompt variants on a subset of the 3{,}000 examples, and it additionally spends calls on generating new candidates (mutations) from the current best prompts. Although it can prune aggressively, the repeated evaluate--expand loop still results in a multi-\(\times\) overhead in calls compared with single-pass induction, consistent with the \(\sim 3\times\) cost in the table.

\item \textbf{OPRO} \cite{conf/iclr/Yang0LLLZC24}. A method that uses LLMs as optimizers by specifying the optimization objective in natural language. In an iterative loop, the LLM proposes new candidates from a meta-prompt containing prior candidates and their scores, progressively improving performance over time. \emph{LLM-call cost:} each OPRO iteration uses (i) one (or a few) optimizer-LLM calls to propose a batch of new instructions and (ii) many evaluation-LLM calls to score those candidates on the 3{,}000 training examples (often via a fixed evaluation subset per iteration). Total calls therefore scale roughly with \#iterations\(\times\)\#candidates\(\times\)\#eval-examples, yielding the moderate budget shown in the table.

\item \textbf{DSPy (MIPROv2)} \cite{conf/emnlp/Opsahl-OngRPBPZ24}. An optimizer for multi-stage language-model programs that jointly tunes free-form instructions to maximize task performance without requiring module-level labels. It uses a grounded proposer to generate high-quality candidates and a Bayesian surrogate model to efficiently search the combinatorial space of prompt configurations. \emph{LLM-call cost:} DSPy amortizes calls by (i) proposing relatively few high-quality candidates per round and (ii) using a surrogate model to prioritize which configurations to evaluate, so it does not need to exhaustively score every candidate on all 3{,}000 datapoints. As a result, the dominant cost is a limited number of evaluation runs (plus occasional proposer calls), leading to the smallest overall call budget in our setting.

\item \textbf{APE} \cite{zhou2023large}. A method that frames instruction generation as natural-language program synthesis: given input--output demonstrations, an LLM generates candidate prompts, which are then scored and filtered to select the instruction that best elicits the desired behavior. \emph{LLM-call cost:} APE requires (i) generation calls to synthesize a pool of candidate instructions from demonstrations and (ii) evaluation calls to score each candidate on a subset (or all) of the 3{,}000 training datapoints. Since the main loop is ``generate a batch, then score and select,'' the total cost is dominated by candidate scoring and scales with \#candidates\(\times\)\#eval-examples.

\end{itemize}

\section{Additional Experimental Details and Results}
\label{app:exp_details}

\subsection{Training Rewards and Hyperparameters}
\label{app:rewards}

Table~\ref{tab:reward_app} specifies the training reward $R(y,y^*)$ used for each task. We use composite rewards when the primary metric is too sparse for stable policy-gradient optimization.

\begin{table}[h]
\centering
\small
\begin{tabular}{ll}
\toprule
\textbf{Task} & \textbf{Training Reward} $R(y,y^*)$ \\
\midrule
Amazon-Beauty & $0.7 \times \mathrm{HR@10} + 0.3 \times \mathrm{NDCG@10}$ \\
HotpotQA      & $0.5 \times \mathrm{F1} + 0.5\times \mathrm{EM}$ \\
SQuAD 2.0     & $0.5 \times \mathrm{F1} + 0.5\times \mathrm{EM}$ \\
PubMedQA      & Accuracy (Yes/No) \\
\bottomrule
\end{tabular}
\caption{Training reward used for each task.}
\label{tab:reward_app}
\end{table}

Unless otherwise specified, we use Adam with learning rate $10^{-4}$, weight decay $10^{-2}$, group size $G=4$, entropy coefficient $\beta_{\mathrm{ent}}=1.0$, and length-regularization weight $\lambda_{\mathrm{len}}=0.01$. The default highlight budget is $\gamma=0.25$ on Amazon-Beauty and $\gamma=0.15$ on the QA tasks. All experiments run on a node with four NVIDIA RTX PRO 6000 GPUs.

\subsection{Span Coalescence and Sampling Strategy}
\label{app:span_ablation}

Table~\ref{tab:span_ablation_app} ablates the span-coalescence gap width $\delta$ and the test-time sampling method on Amazon-Beauty. Performance is stable across a broad range of widths and sampling strategies.

\begin{table}[h]
\centering
\small
\begin{tabular}{lcccccc}
\toprule
\textbf{Width} & \multicolumn{2}{c}{\textbf{Greedy-TopK}} & \multicolumn{2}{c}{\textbf{Softmax}} & \multicolumn{2}{c}{\textbf{Gumbel-TopK}} \\
\cmidrule(lr){2-3}\cmidrule(lr){4-5}\cmidrule(lr){6-7}
& HR@10 & NDCG@10 & HR@10 & NDCG@10 & HR@10 & NDCG@10 \\
\midrule
4  & 0.0271 & 0.0150 & 0.0271 & 0.0147 & 0.0266 & 0.0147 \\
6  & 0.0270 & 0.0152 & 0.0273 & 0.0150 & 0.0270 & 0.0152 \\
8  & 0.0283 & 0.0157 & 0.0277 & 0.0153 & 0.0273 & 0.0152 \\
10 & 0.0288 & 0.0159 & 0.0282 & 0.0155 & 0.0270 & 0.0150 \\
12 & 0.0281 & 0.0158 & 0.0260 & 0.0149 & 0.0275 & 0.0153 \\
14 & 0.0271 & 0.0154 & 0.0267 & 0.0150 & 0.0283 & 0.0152 \\
16 & 0.0273 & 0.0155 & 0.0266 & 0.0149 & 0.0276 & 0.0154 \\
\bottomrule
\end{tabular}
\caption{Span-coalescence width and sampling-method ablation on Amazon-Beauty.}
\label{tab:span_ablation_app}
\end{table}

\subsection{Loss Component Ablation}
\label{app:loss_ablation}

Table~\ref{tab:loss_ablation_app} ablates the length-regularization weight $\lambda_{\mathrm{len}}$ and entropy coefficient $\beta_{\mathrm{ent}}$. Removing either regularizer degrades performance, suggesting that they play complementary roles: entropy maintains exploration, while length regularization stabilizes expected selection density.

\begin{table}[h]
\centering
\small
\begin{tabular}{lccc}
\toprule
$\lambda_{\mathrm{len}}$ & $\beta_{\mathrm{ent}}$ & HR@10 & NDCG@10 \\
\midrule
0.0   & 1.0 & 0.0267 & 0.0149 \\
0.001 & 1.0 & 0.0271 & 0.0153 \\
0.005 & 1.0 & 0.0284 & 0.0152 \\
0.01  & 1.0 & 0.0287 & 0.0157 \\
0.01  & 0.5 & 0.0276 & 0.0153 \\
0.01  & 0.1 & 0.0278 & 0.0156 \\
0.01  & 0.0 & 0.0268 & 0.0149 \\
\bottomrule
\end{tabular}
\caption{Loss-coefficient ablation on Amazon-Beauty.}
\label{tab:loss_ablation_app}
\end{table}

\subsection{End-to-end Latency Breakdown}\label{app:latency}
Table~\ref{tab:latency} reports wall-clock Actor latency across model sizes. Even the 4B Actor used in our most demanding setting adds under 0.34\,s (p95), compared with 8--18\,s for the 14B Solver. The total inference-time overhead of the Actor is therefore roughly 1.3\%--2.9\% of the pipeline latency. Our latency measurements are for \emph{per-query} inference, where each query induces one Actor pass and one Solver call. Conversational multi-turn serving with cache-aware reuse across changing query-conditioned highlights would require an additional systems layer and is left to future work.

\begin{table}[h]
\centering
\small
\begin{tabular}{lccccc}
\toprule
\textbf{Latency (s)} & \textbf{Actor 0.6B} & \textbf{Actor 1.7B} & \textbf{Actor 4B} & \textbf{Actor 8B} & \textbf{Solver 14B} \\
\midrule
mean & 0.047 & 0.102 & 0.233 & 0.528  & 8.077 to 18.403 \\
p50  & 0.051 & 0.123 & 0.237 & 0.546 & 8.086 to 19.155 \\
p95  & 0.057 & 0.130 & 0.336 & 0.635 & 9.085 to 19.589 \\
\bottomrule
\end{tabular}
\caption{Actor latency across model sizes. Solver latency varies by task.}
\label{tab:latency}
\end{table}

\subsection{Sensitivity Analysis}\label{app:sensitivity}

\begin{figure}[h!]
    \centering
    \includegraphics[width=0.55\columnwidth]{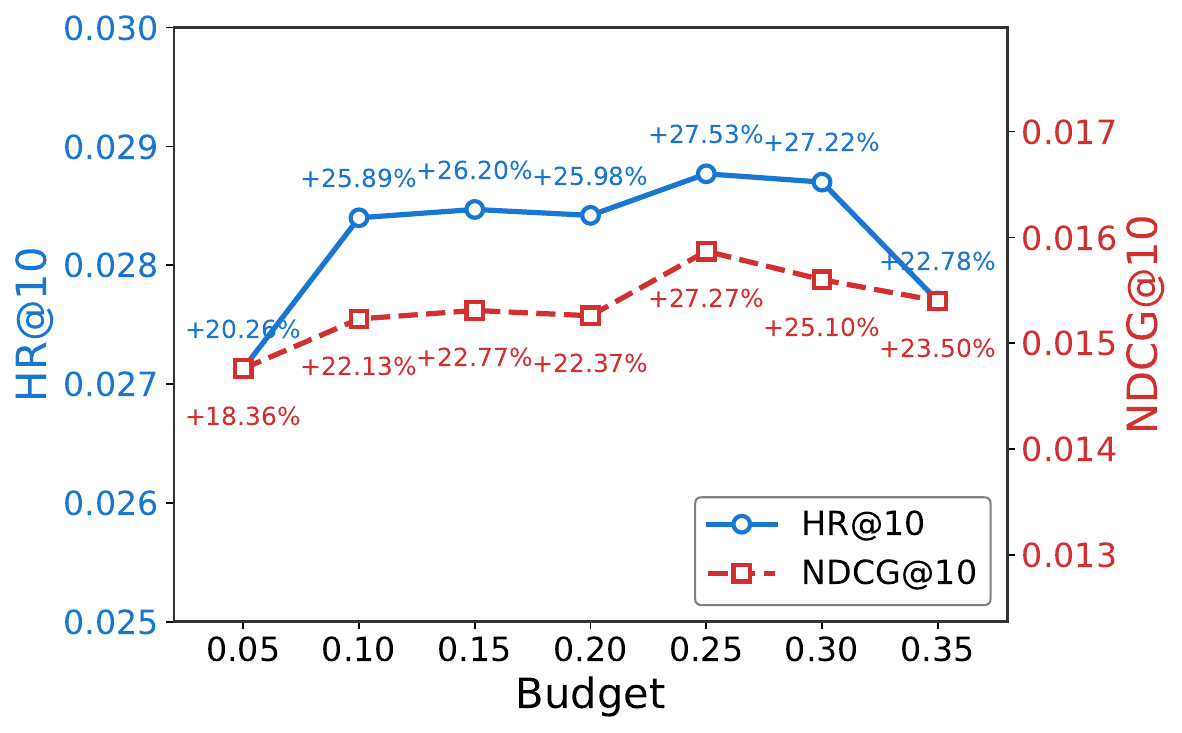} 
    \caption{Sensitivity to highlight budget $\gamma$ on Amazon-Beauty. Percentages denote improvement over MI.}
    \label{fig:tease}
\end{figure}

\textbf{Sensitivity to highlight budget.}
We vary the highlight budget $\gamma$, which caps the fraction of tokens the Actor may emphasize, and report performance on Amazon-Beauty in Figure~\ref{fig:tease}. Two trends emerge. First, \HiLight delivers gains even at small budgets (e.g., $\gamma\le 0.10$), suggesting that emphasizing a small number of spans can already improve ranking in long histories. Second, performance is broadly stable over a wide mid-range ($\gamma \in [0.10,0.30]$), with a mild optimum around $\gamma \approx 0.25$. This plateau indicates that \HiLight is not brittle to $\gamma$: once the budget is sufficient to cover the key spans, additional highlighting yields diminishing returns and may slightly reduce contrast between emphasized evidence and distractors. \textit{Practically, this means \HiLight can be deployed without extensive per-task budget tuning.}

\textbf{Sensitivity to highlight markers.}
We test whether \HiLight’s gains depend on the specific tag tokenization used to denote emphasis. Concretely, we \emph{train a single Actor using the default marker format} (the one used in Table~\ref{tab:overall}) and, at inference time, \emph{swap the marker syntax} while keeping the selected spans, \rev{checkpoint, context, decoding settings,} and highlight budget fixed. This isolates marker-tokenization effects from span-selection quality, since the Actor and selected spans are unchanged. The emphasized text content is identical; only the boundary tokens that denote emphasis are replaced. Table~\ref{tab:ablation_tag_full} shows that performance varies only modestly across Markdown-style, symbol-delimited, and HTML/XML-style markers: while the default format remains best, alternative syntaxes retain most of the gain. This indicates the learned policy is not overfitting to a particular delimiter tokenization.

\begin{table}[h]
\centering
\small
\setlength{\tabcolsep}{10pt} 
\begin{tabular}{lcc}
\toprule
\textbf{Variant} & \textbf{HR@10} & \textbf{NDCG@10} \\
\midrule
\texttt{**...**}  & 0.02837 & 0.01512 \\
\texttt{[[...]]}  & 0.02852 & 0.01547 \\
\texttt{\{...\}}  & 0.02852 & 0.01547 \\
\texttt{>>...<<}  & 0.02853 & 0.01512 \\
\texttt{<b>...</b>}  & 0.02852 & 0.01575 \\
\texttt{<important>} & 0.02854 & 0.01570 \\
\midrule
\textbf{Default} & 0.02877 & 0.01587 \\
\bottomrule
\end{tabular}
\caption{Marker-format robustness on Amazon-Beauty. We train \HiLight with the \textbf{default} marker format (Table~\ref{tab:overall}) and replace \emph{only} boundary-marker tokens at inference time (selected spans unchanged). Performance is stable across Markdown, symbol, and HTML/XML-style syntaxes.}
\label{tab:ablation_tag_full}
\end{table}

\begin{revblock}
\textbf{Sensitivity to Solver prompt templates.}
We further vary the Solver prompt on HotpotQA using five templates: the default instruction used in the paper; two independent paraphrases with identical content, the second in a more verbose register; a minimal template that keeps only the bare answer-format contract; and an order-swapped template that uses the default instruction but places the evidence before the question. Table~\ref{tab:prompt_templates} shows that \HiLight yields higher EM and F1 under all five templates, with relative improvements of 4.23\% to 9.29\% in EM and 3.61\% to 7.67\% in F1. The minimal template lowers absolute performance for both conditions, suggesting that prompt quality and evidence emphasis provide complementary benefits.
\end{revblock}

\begin{table}[h]
\color{\revcolor}
\centering
\small
\begin{tabular}{lcccc}
\toprule
& \multicolumn{2}{c}{\textbf{MI}} & \multicolumn{2}{c}{\textbf{\HiLight}} \\
\cmidrule(lr){2-3}\cmidrule(lr){4-5}
\textbf{Template} & \textbf{EM} & \textbf{F1} & \textbf{EM} & \textbf{F1} \\
\midrule
Default       & 0.556 & 0.706 & 0.606 (+8.99\%) & 0.741 (+4.96\%) \\
Paraphrase A  & 0.574 & 0.711 & 0.606 (+5.57\%) & 0.749 (+5.34\%) \\
Paraphrase B  & 0.554 & 0.690 & 0.592 (+6.86\%) & 0.736 (+6.67\%) \\
Order swapped & 0.560 & 0.704 & 0.612 (+9.29\%) & 0.758 (+7.67\%) \\
Minimal       & 0.426 & 0.581 & 0.444 (+4.23\%) & 0.602 (+3.61\%) \\
\bottomrule
\end{tabular}
\caption{\rev{Solver prompt-template sensitivity on HotpotQA. Percentages are relative improvements of \HiLight over MI under the same template.}}
\label{tab:prompt_templates}
\end{table}

\textbf{Sensitivity to Actor capacity.}
We fix the Solver at Qwen3-14B and vary Actor size from 0.6B to 8B. Table~\ref{tab:actor_size} shows that Actor scaling helps most on harder evidence-selection tasks (HotpotQA), while gains are smaller on tasks where evidence is more localized (Amazon-Beauty). Returns diminish beyond mid-sized Actors (around 4B). In practice, this means that Actor sizing can be guided by the structural complexity of the evidence-selection problem rather than the raw context length.

\begin{table}[h]
\centering
\small
\begin{tabular}{lcccc}
\toprule
\multirow{2}{*}{\textbf{Actor}} & \multicolumn{2}{c}{\textbf{Amazon-Beauty}} & \multicolumn{2}{c}{\textbf{HotpotQA}} \\
\cmidrule(lr){2-3}\cmidrule(lr){4-5}
& HR@10 & NDCG@10 & EM & F1\\
\midrule
0.6B & 0.02830 & 0.01521 & 0.593 & 0.721 \\
1.7B & 0.02877 & 0.01587 & 0.597 & 0.728  \\
4B   & 0.02937 & 0.01620 & 0.606 & 0.741\\
8B   & 0.02994 & 0.01648 & 0.609 & 0.742 \\
\bottomrule
\end{tabular}
\caption{Actor size vs.\ performance. Solver = Qwen3-14B.}
\label{tab:actor_size}
\end{table}

\textbf{Sensitivity to context length.}
To test \HiLight under genuinely long contexts, we inject distractor documents into HotpotQA to extend input length from the original setting to approximately 8K, 16K, and 32K tokens.  Table~\ref{tab:longctx} shows that \HiLight maintains its advantage even as context length and distractor load increase. While absolute gains decrease slightly at longer contexts, the relative improvement remains positive throughout, consistent with our hypothesis that \HiLight targets evidence sparsity under distractors rather than absolute context length alone.

\begin{table}[h]
\centering
\small
\begin{tabular}{lcccc}
\toprule
\multirow{2}{*}{\textbf{Method}} & \multicolumn{4}{c}{\textbf{Context Length}} \\
\cmidrule(lr){2-5}
& $\sim$2K (original) & $\sim$8K & $\sim$16K & $\sim$32K \\
\midrule
MI (EM)      & 0.556 & 0.524 & 0.518 & 0.512 \\
\HiLight (EM) & 0.606 & 0.550 & 0.542 & 0.532 \\
$\Delta_{\%}$     & +9.0\% & +5.0\% & +4.6\% & +3.9\% \\
\midrule
MI (F1)      & 0.706 & 0.683 & 0.675 & 0.656 \\
\HiLight (F1) & 0.741 & 0.710 & 0.693 & 0.682 \\
$\Delta_{\%}$     & +5.0\% & +4.0\% & +2.7\% & +4.0\% \\
\bottomrule
\end{tabular}
\caption{HotpotQA performance at increasing context lengths. Distractor documents injected to extend context.}
\label{tab:longctx}
\end{table}

\begin{revblock}
\textbf{Evidence selection and Actor cost versus context length.}
To examine how the Actor's evidence selection scales with length, we extend the first 300 examples of the HotpotQA validation split to approximately 8K, 16K, and 32K Actor tokens in two ways, holding fixed the 4B Actor, the frozen Qwen3-14B Solver, the document-order shuffling procedure, and a selection budget of $k=4$ sentences. \emph{Injected distractors} append paragraphs drawn from other validation questions as additional documents. \emph{Wikipedia extensions} retain every original HotpotQA paragraph verbatim and extend it with the continuation of its Wikipedia article, removing the introduction to avoid duplication and dropping reference-style sections; if the source articles are exhausted before the target length, additional complete Wikipedia articles are appended as standalone documents. Retaining the original paragraphs verbatim keeps the annotated supporting sentences traceable. Supporting-fact recall@$k$ is the fraction of annotated supporting sentences covered by the top four selected sentences; zero-gold failure is the fraction of examples for which no selected sentence is annotated as supporting evidence; and both-hop coverage is the fraction of examples whose selection contains at least one supporting sentence from each of the two gold paragraphs.

Table~\ref{tab:longctx_select} shows that Wikipedia extensions yield lower supporting-fact recall and both-hop coverage than injected distractors at all three lengths. From 8K to 32K tokens, recall decreases from 0.755 to 0.689 with injected distractors and from 0.706 to 0.666 with Wikipedia extensions. At equal target lengths, the two settings have the same Actor resource profile: mean Actor latency is 0.41, 0.95, and 2.35 seconds at 8K, 16K, and 32K tokens, compared with 0.23 seconds at the original length of about 2K tokens, and peak incremental GPU memory above the post-loading baseline is 4.4, 8.5, and 17.2 GB, compared with 0.8 GB. All measurements were obtained with the Actor and Solver resident on a single 96 GB GPU, and no out-of-memory failures occurred.
\end{revblock}

\begin{table}[h]
\color{\revcolor}
\centering
\small
\begin{tabular}{llccc}
\toprule
\textbf{Context} & \textbf{Extension} & \textbf{Recall@$k$} & \textbf{Zero-gold failure} & \textbf{Both-hop coverage} \\
\midrule
8K  & Injected distractors & 0.755 & 1.3\% & 0.593 \\
8K  & Wikipedia extensions & 0.706 & 5.0\% & 0.537 \\
\midrule
16K & Injected distractors & 0.739 & 1.3\% & 0.577 \\
16K & Wikipedia extensions & 0.681 & 6.3\% & 0.490 \\
\midrule
32K & Injected distractors & 0.689 & 5.7\% & 0.507 \\
32K & Wikipedia extensions & 0.666 & 5.7\% & 0.470 \\
\bottomrule
\end{tabular}
\caption{\rev{Evidence selection of the 4B Actor on 300 HotpotQA validation examples extended to increasing Actor-token lengths, with a selection budget of $k=4$ sentences.}}
\label{tab:longctx_select}
\end{table}

\subsection{Transfer to API-Based Solvers}\label{app:api_solver}

To validate practical black-box compatibility beyond open-weight models, we apply the same Actor trained with Qwen3-14B zero-shot to GPT-5 mini via its API. Table~\ref{tab:gpt5mini} shows that \HiLight yields positive gains on both Amazon-Beauty and HotpotQA. The gains are more modest than in high-distractor recommendation settings, which is expected because the target Solver is substantially stronger and may already resolve more of the context on its own. Nevertheless, the result is important: \HiLight can be deployed with an API-only Solver at inference time, without Solver gradients, logits, or internal activations.

\begin{table}[h]
\centering
\small
\begin{tabular}{lcccc}
\toprule
\multirow{2}{*}{\textbf{Solver}} & \multicolumn{2}{c}{\textbf{Amazon-Beauty}} & \multicolumn{2}{c}{\textbf{HotpotQA}} \\
\cmidrule(lr){2-3}\cmidrule(lr){4-5}
& HR@10 & NDCG@10 & EM & F1 \\
\midrule
GPT-5 mini (MI)       & 0.03083 & 0.01941 & 0.646 & 0.809 \\
GPT-5 mini + \HiLight & 0.03127 & 0.02070 & 0.657 & 0.818 \\
$\Delta_{\%}$         & +1.43\% & +6.65\% & +1.70\% & +1.11\% \\
\bottomrule
\end{tabular}
\caption{GPT-5 mini evaluation. Actor trained on Qwen3-14B and applied zero-shot.}
\label{tab:gpt5mini}
\end{table}

\begin{revblock}
\textbf{Additional API-based Solvers.}
We further apply the same Actor without adaptation to five additional API-based Solver configurations on HotpotQA: GPT-5 (low), GPT-5 (high), Claude Haiku 4.5, Claude Sonnet 5, and Claude Opus 4.8. Table~\ref{tab:api_solvers} in Section~\ref{subsec:transfer} reports relative improvements over each target Solver's own MI baseline, together with GPT-5 mini from Table~\ref{tab:gpt5mini}. Both EM and F1 improve for all six configurations. Within each family, the gains are smaller for stronger Solvers, consistent with stronger Solvers already resolving more of the context on their own; confirming this trend statistically would require uncertainty estimates for the paired differences.
\end{revblock}

\subsection{Statistic Confidence}
\label{app:stat}
To assess the statistical reliability of the observed improvements, we report 95\% confidence intervals computed via a paired bootstrap procedure. For each comparison between HiLight and a baseline $B$, we draw $R = 1000$ bootstrap replicates. In each replicate, we sample $n = 1000$ example indices uniformly with replacement from the test set and apply the \emph{same} sampled indices to both methods before computing the paired relative improvement
\[
\Delta^{\%}_b = 100 \cdot \frac{\mathrm{Metric}_{\text{HiLight}}(S_b) - \mathrm{Metric}_{B}(S_b)}{\mathrm{Metric}_{B}(S_b)},
\]
where $S_b$ is the sampled subset. The 95\% confidence interval is reported as the percentile interval
\[
\left[\mathrm{percentile}_{2.5}\{\Delta^{\%}_b\}_{b=1}^{R},\ \mathrm{percentile}_{97.5}\{\Delta^{\%}_b\}_{b=1}^{R}\right].
\]
 Pairing is essential here: because both methods are evaluated on the same examples, jointly resampling indices removes the variance attributable to example difficulty and isolates the method-level effect, yielding tighter and more appropriate intervals than independently bootstrapping each method. We apply this procedure to all reported metrics (HR@10, NDCG@10, EM, and token-F1).

\begin{table}[h]
\centering
\small
\setlength{\tabcolsep}{6pt}
\begin{tabular}{ll cc}
\toprule
Dataset & Metric & $\Delta^{\%}$ (\HiLight vs.\ Best Base.) & $\Delta^{\%}$ (\HiLight vs.\ MI) \\
\midrule
\multirow{2}{*}{Amazon-Beauty}
  & HR@10   & $+4.63\%$ {\scriptsize $[+0.59, +8.83]$}   & $+27.03\%$ {\scriptsize $[+20.07, +34.47]$} \\
  & NDCG@10 & $+10.74\%$ {\scriptsize $[+6.42, +14.02]$} & $+26.00\%$ {\scriptsize $[+19.20, +33.44]$} \\
\midrule
\multirow{2}{*}{HotpotQA}
  & EM & $+2.12\%$ {\scriptsize $[+0.60, +4.05]$} & $+8.69\%$ {\scriptsize $[+5.71, +10.78]$} \\
  & F1 & $+2.58\%$ {\scriptsize $[+1.02, +4.17]$} & $+4.97\%$ {\scriptsize $[+2.13, +6.89]$} \\
\midrule
\multirow{2}{*}{SQuAD 2.0}
  & EM & $+2.39\%$ {\scriptsize $[+1.40, +4.08]$}  & $+12.62\%$ {\scriptsize $[+8.54, +14.65]$}  \\
  & F1 & $+4.46\%$ {\scriptsize $[+2.87, +6.14]$}  & $+10.95\%$ {\scriptsize $[+7.02, +14.05]$}  \\
\midrule
\multirow{2}{*}{PubMedQA}
  & Acc.  & $+1.27\%$ {\scriptsize $[+0.36, +2.52]$} & $+1.87\%$ {\scriptsize $[+1.07, +2.75]$}\\
  & M-F1  & $+5.42\%$ {\scriptsize $[+3.78, +6.97]$} & $+5.42\%$ {\scriptsize $[+3.78, +6.97]$} \\
\bottomrule
\end{tabular}
\caption{Relative improvement $\Delta^{\%}$ of \textsc{HiLight} over the strongest baseline and over MI. Intervals that exclude $0$ indicate statistically significant improvement at the 95\% level.}
\label{tab:bootstrap_ci}
\end{table}

\begin{table}[h]
\centering
\small
\setlength{\tabcolsep}{6pt}
\begin{tabular}{ll c}
\toprule
\textbf{Model} & \textbf{Metric} & $\Delta^{\%}$ (\HiLight vs. MI) \\
\midrule
\multirow{2}{*}{Qwen3-4B}
  & HR@10   & $+13.87\%$ {\scriptsize $[+7.89, +20.63]$} \\
  & NDCG@10 & $+7.93\%$  {\scriptsize $[+3.41, +13.07]$} \\
\midrule
\multirow{2}{*}{Qwen3-8B}
  & HR@10   & $+13.34\%$ {\scriptsize $[+6.73, +19.21]$} \\
  & NDCG@10 & $+12.26\%$ {\scriptsize $[+5.93, +18.08]$} \\
\midrule
\multirow{2}{*}{Gemma-3-27B}
  & HR@10   & $+15.74\%$ {\scriptsize $[+9.21, +23.46]$} \\
  & NDCG@10 & $+29.87\%$ {\scriptsize $[+20.84, +40.63]$} \\
\midrule
\multirow{2}{*}{Qwen3-32B}
  & HR@10   & $+7.83\%$  {\scriptsize $[+1.97, +13.52]$} \\
  & NDCG@10 & $+19.24\%$ {\scriptsize $[+10.83, +30.38]$} \\
\midrule
\multirow{2}{*}{Llama-3-70B}
  & HR@10   & $+18.43\%$ {\scriptsize $[+10.67, +25.89]$} \\
  & NDCG@10 & $+26.13\%$ {\scriptsize $[+16.97, +34.84]$} \\
\bottomrule
\end{tabular}
\caption{Transfer of \textsc{HiLight}'s emphasis actor (trained on Qwen3-14B) to smaller and larger solver LLMs on Amazon-Beauty.}
\label{tab:transfer_stat}
\end{table}

\begin{table}[h]
\centering
\small
\setlength{\tabcolsep}{6pt}
\begin{tabular}{ll c}
\toprule
\textbf{Dataset} & \textbf{Metric} & $\Delta^{\%}$ (\HiLight vs.\ MI) \\
\midrule
\multirow{2}{*}{Amazon-Beauty}
  & HR@10   & $+1.31\%$ {\scriptsize $[+0.19, +2.87]$} \\
  & NDCG@10 & $+6.14\%$ {\scriptsize $[+0.83, +10.97]$} \\
\midrule
\multirow{2}{*}{HotpotQA}
  & EM & $+1.41\%$ {\scriptsize $[+0.45, +2.13]$} \\
  & F1 & $+1.04\%$ {\scriptsize $[+0.18, +1.79]$} \\
\bottomrule
\end{tabular}
\caption{Zero-shot transfer of \textsc{HiLight}'s emphasis actor (trained on Qwen3-14B) to GPT-5 mini as the solver .}
\label{tab:gpt5mini_stat}
\end{table}
\subsection{Interpretability \& Visualization}
\label{app:case}

\paragraph{Interpretability.}
Figure~\ref{fig:case_study_box} illustrates a representative Amazon-Beauty case study where \HiLight effectively steers the frozen Solver through high-noise user history. While the baseline model is distracted by irrelevant interactions and ranks the ground truth item 14th , our Actor successfully isolates decisive signals by highlighting explicit feature alignment (e.g., a user's desire for ``grips'') and strong sentiment. These structural markers create a semantic bridge to the target product, improving its rank to 5th.
This example underscores the \textit{auditability} of our framework: unlike black-box soft attention, the highlighted spans provide a human-readable rationale (e.g., the matching of ``grips'') that explains \textit{why} the model improved its prediction.

\begin{figure*}[h]
\centering
\begin{tcolorbox}[ 
    colframe=blue!57,
    boxrule=0.5pt,
    arc=2mm,
    width=\textwidth,
    title=\textbf{Case Study: Amazon-Beauty Item Re-ranking}
]
\small
\textbf{Query:}\\
Given a user's history summary and a list of candidate items, re-rank the candidates according to their likelihood of being the next item of interest.

\vspace{0.2cm}
\hrule
\vspace{0.2cm}

\textbf{Context:}\\
...title=``Great brush cleaner", text=``Great brush cleaner! Easy to use \hl{but it would be better if it had grips on the back} so it doesn't move around the sink" \\
...title=``I am in LOVE with these", text=``\hl{ I am in LOVE with these! I would buy again.} They're easy to pack because ..."

\vspace{0.2cm}
\hrule
\vspace{0.2cm}

\textbf{Comparison \& Outcome:}\\
$\bullet$ Standard Baseline Output: Ranked \textbf{14th} (Failed to retrieve)

$\bullet$ Ours Output: Ranked \textbf{5th} (Successfully improved ranking)

$\bullet$ Ground Truth Item $y^*$: e.l.f. Tone Adjusting Face Primer ... {\textbf{Grips Makeup To Last}}, Veg

\end{tcolorbox}
\caption{A case study from the Amazon-Beauty dataset, with \hl{purple} text emphasizing the evidence. \HiLight successfully identifies the user's specific preference for ``grips", allowing the frozen Solver to match it with the target product characteristic.}
\label{fig:case_study_box}
\end{figure*}

\textbf{Visualization of actor scores.}
To better understand the Actor's selection behavior beyond discrete highlight tags, we visualize token-level importance scores on a representative HotpotQA instance in Figure~\ref{fig:viz}. We observe a small number of pronounced peaks: the first aligns with the query tokens, suggesting the Actor conditions on the question before scoring the context, and the remaining peaks concentrate on a few context regions that substantially overlap with the annotated supporting evidence (red spans). Between these peaks, scores remain low, indicating that the Actor focuses its budget on a limited set of candidate evidence spans rather than broadly reweighting the full input. While this visualization is qualitative, the peaked structure is consistent with \HiLight’s intended role as a budgeted evidence selector for multi-hop reasoning.

\textbf{Quantitative evidence alignment.}
To move beyond qualitative examples, we quantify the overlap between Actor highlights and ground-truth supporting facts annotated in HotpotQA. We threshold Actor scores at 0.5 to obtain predicted evidence tokens and compute precision, recall, and F1 against the annotations. Table~\ref{tab:evidence_overlap} shows that the Actor achieves up to 0.78 F1 overlap with human evidence, despite never seeing evidence labels during training. This suggests that optimizing for task reward alone can implicitly recover meaningful evidence structure, providing a form of weak supervision for evidence selection without requiring explicit annotations. Comparing with the Actor-size study (Table~\ref{tab:actor_size}), the monotonic improvement in evidence F1 from 0.68 (0.6B) to 0.78 (8B) mirrors the downstream task gains, confirming that additional Actor capacity translates into genuinely better evidence identification rather than exploiting spurious patterns.

\textbf{Failure analysis.}
We analyze cases where the Solver is correct under MI but becomes incorrect after applying \HiLight. Table~\ref{tab:failure} shows the breakdown. For Amazon-Beauty, we randomly sample 1000 test points whose ground truth appears in the candidate set (examples without ground truth are filtered out); for HotpotQA, we sample 1000 points from its 7405-example test set. On HotpotQA, where annotated supporting facts enable fine-grained diagnosis, 16.5\% of failures are due to misleading emphasis on non-decisive spans, while 33.8\% occur when the Actor surfaces relevant evidence but the Solver still fails at downstream reasoning. Across both tasks, the number of newly correct predictions (success cases) substantially exceeds the number of newly incorrect ones (failure cases).

\begin{table}[h]
\centering
\small
\begin{tabular}{lcc}
\toprule
\textbf{Failure Mode} & \textbf{Amazon-Beauty} & \textbf{HotpotQA} \\
\midrule
Misleading emphasis on non-decisive spans & N/A & 16.5\% \\
Evidence surfaced, but Solver still fails & N/A  &  33.8\%\\
\midrule
Total success (MI incorrect $\rightarrow$ \HiLight correct) & 168 & 63 \\
Total failure (MI correct $\rightarrow$ \HiLight incorrect) & 28 & 20 \\
Total test samples & 1000 & 1000 \\
\bottomrule
\end{tabular}
\caption{Failure-case breakdown for instances where \HiLight changes a correct MI prediction into an incorrect one.}
\label{tab:failure}
\end{table}

\begin{revblock}
\subsection{Matched Evidence-Selection Controls}\label{app:controls}

Table~\ref{tab:controls} complements the fixed-Actor Pruned ablation in Table~\ref{tab:ablation} with additional HotpotQA controls. All controls use the same frozen Qwen3-14B Solver and evaluation set as \HiLight. The trained pruning selector and the trained LLM highlighter also use the same training prefix, downstream $0.5\,\mathrm{EM}+0.5\,\mathrm{F1}$ reward, 40K Solver-query budget, and checkpoint-selection rule, thereby matching data access, supervision, shared Solver cost, and model selection.

\textbf{Trained pruning selector.} This control uses the same Actor architecture and GRPO procedure as \HiLight, but places the pruning operator inside the RL loop, so that rewards are computed from the selected sentences alone.

\textbf{Trained LLM highlighter.} An LLM matching the \HiLight Actor in size generates the IDs of sentences to highlight and is optimized with sequence-level GRPO ($G=4$, with group-normalized advantages) using the same downstream reward. Its selections are rendered with \HiLight's marker operator. Its full-context prefill and autoregressive decoding give a different Actor-side cost profile from \HiLight's single token-scoring pass, while the supervision and the Solver-query budget remain matched.

\textbf{Prompt-tuned Solver.} We apply PEFT prompt tuning to the frozen Solver (bf16) by prepending 32 virtual tokens ($\approx$164K trainable parameters) whose embeddings are initialized from a system prompt. We train with AdamW (learning rate $5\times10^{-3}$, batch size 8) for one epoch on 10,000 HotpotQA examples, applying cross-entropy loss only to the gold-answer target, and evaluate with greedy decoding on the HotpotQA distractor validation split, which is disjoint from the prompt-tuning examples. Unlike \HiLight, this diagnostic requires gradient access to the Solver.

\textbf{Off-the-shelf reranker.} \texttt{bge-reranker-v2-m3} scores every question and sentence pair, and the top-$k$ sentences are passed to the Solver without HotpotQA-specific adaptation.

\textbf{GPT-5 mini API highlighter.} GPT-5 mini selects up to $k$ sentence IDs from the numbered context in a zero-shot setting. The selected sentences are rendered with the same marker operator, and the same frozen Qwen3-14B Solver produces the answer. This control requires one proprietary API call per example.

\begin{table}[h]
\color{\revcolor}
\centering
\small
\begin{tabular}{lcc}
\toprule
\textbf{Method} & \textbf{EM} & \textbf{F1} \\
\midrule
MI & 0.556 & 0.706 \\
Prompt-tuned Solver & 0.587 & 0.727 \\
Trained pruning selector & 0.563 & 0.707 \\
Trained LLM highlighter & 0.572 & 0.712 \\
Off-the-shelf reranker & 0.568 & 0.710 \\
\midrule
GPT-5 mini API highlighter & 0.612 & 0.766 \\
\midrule
\HiLight & 0.606 & 0.741 \\
\bottomrule
\end{tabular}
\caption{\rev{HotpotQA comparison with additional evidence-selection controls. All methods use the same frozen Qwen3-14B Solver and evaluation set; the trained pruning selector, the trained LLM highlighter, and \HiLight share the optimization pool, downstream reward, 40K Solver-query budget, and checkpoint-selection rule.}}
\label{tab:controls}
\end{table}
\end{revblock}

\begin{revblock}
\subsection{Cross-Task Transfer}\label{app:cross_task}

To test whether an Actor trained on one task remains useful on a related unseen task, we apply the HotpotQA Actor unchanged to SQuAD 2.0, without target-task fine-tuning or model selection. Table~\ref{tab:cross_task} reports changes relative to the MI baseline of the evaluation task under its native evaluation protocol. The transferred Actor improves EM by +3.58\% and F1 by +4.64\%, below the +12.80\% and +11.44\% of the task-specific SQuAD 2.0 Actor, indicating that an Actor trained on one QA task retains useful evidence-selection behavior on a related task.

\begin{table}[h]
\color{\revcolor}
\centering
\small
\begin{tabular}{lllcc}
\toprule
\textbf{Actor training task} & \textbf{Evaluation task} & \textbf{Setting} & $\Delta_{\%}$\textbf{EM} & $\Delta_{\%}$\textbf{F1} \\
\midrule
SQuAD 2.0 & SQuAD 2.0 & in-task reference    & +12.80\% & +11.44\% \\
HotpotQA  & SQuAD 2.0 & zero-shot cross-task & +3.58\%  & +4.64\% \\
\bottomrule
\end{tabular}
\caption{\rev{Cross-task transfer to SQuAD 2.0 ($\Delta_{\%}$ over the SQuAD 2.0 MI baseline).}}
\label{tab:cross_task}
\end{table}
\end{revblock}

\section{Additional Case Studies}
\label{app:case_studies}

We provide additional qualitative case studies on HotpotQA and PubMedQA to illustrate how evidence emphasis
mitigates common long-context failure modes, such as entity-level confusion and evidence dilution,
thereby improving the frozen Solver's final prediction. 
\begin{figure*}[h!]
\centering
\begin{tcolorbox}[
    colframe=blue!57,
    boxrule=0.5pt,
    arc=2mm,
    width=\textwidth,
    title=\textbf{Case Study: HotpotQA Multi-hop QA}
]
\small
\textbf{Query:}\\
According to the 2001 census, what was the population of the city in which Kirton End is located?

\vspace{0.2cm}
\hrule
\vspace{0.2cm}

\textbf{Context:}\\
...Boston is a town and small port in Lincolnshire, on the east coast of England... \\
...The town itself had \hl{a population of 35,124 at the 2001 census}... \\
...\hl{Kirton End is a hamlet in the civil parish of Kirton in the Boston district of Lincolnshire, England}...

\vspace{0.2cm}
\hrule
\vspace{0.2cm}

\textbf{Comparison \& Outcome:}\\
$\bullet$ Standard Baseline Output: \textbf{341} (Incorrect; confused with the hamlet/village population)\\
$\bullet$ Ours Output: \textbf{35,124} (Correct)\\
$\bullet$ Ground Truth $y^*$: \textbf{35,124}

\end{tcolorbox}
\caption{A case study from the HotpotQA dataset, with \hl{purple} text emphasizing the decisive evidence. \HiLight highlights (i) that Kirton End is in the Boston district and (ii) Boston's 2001 census population, enabling the frozen Solver to answer correctly.}
\label{fig:case_hotpot_box}
\end{figure*}

\begin{figure*}[h!]
\centering
\begin{tcolorbox}[
    colframe=blue!57,
    boxrule=0.5pt,
    arc=2mm,
    width=\textwidth,
    title=\textbf{Case Study: PubMedQA Biomedical Yes/No QA}
]
\small
\textbf{Query:}\\
Do wastewater treatment plants release large amounts of extended-spectrum $\beta$-lactamase-producing \textit{Escherichia coli} into the environment?

\vspace{0.2cm}
\hrule
\vspace{0.2cm}

\textbf{Context:}\\
...The determinants of the spread of extended-spectrum $\beta$-lactamase-producing \textit{E. coli} (ESBLEC) in the community remain unclear... \\
...\hl{we analyzed the ESBLEC population throughout an urban wastewater network}... \\
...\hl{The \textit{E. coli} load was higher in urban wastewater than in hospital wastewater}...

\vspace{0.2cm}
\hrule
\vspace{0.2cm}

\textbf{Comparison \& Outcome:}\\
$\bullet$ Standard Baseline Output: \textbf{No} (Incorrect)\\
$\bullet$ Ours Output: \textbf{Yes} (Correct)\\
$\bullet$ Ground Truth $y^*$: \textbf{Yes}

\end{tcolorbox}
\caption{A case study from the PubMedQA dataset, with \hl{purple} text emphasizing the key evidence. By highlighting the study's findings on ESBLEC prevalence and \textit{E. coli} load in wastewater, \HiLight helps the frozen Solver infer the correct ``Yes'' answer.}
\label{fig:case_pubmed_box}
\end{figure*}

\newpage

\section{Reproducibility}
\label{app:configs}

\subsection{Actor Model Architecture}
\label{sec:actor_arch}

\paragraph{Backbone LLM.}
We implement the actor on top of a pretrained causal language model (Qwen family). Given an input sequence \(\{x_i\}_{i=1}^{L}\) (token IDs) and an attention mask, we run a standard forward pass to obtain per-layer hidden representations \(\{\mathbf{h}_i^{(\ell)}\}\). To reduce overhead and keep the actor purely as a feature extractor, we disable KV caching.
\paragraph{Freezing the backbone.}
During actor training, we freeze the pretrained backbone parameters and prevent gradient updates to the full LLM. Concretely, we set \(\nabla_{\theta_{\text{LM}}}=0\), so the LLM serves as a fixed encoder. This substantially reduces the number of trainable parameters and improves training stability.
\paragraph{LoRA-based parameter-efficient adaptation.}
To enable task-specific adaptation without updating the full backbone, we optionally attach LoRA adapters. We adopt a standard LoRA configuration with rank \(r=8\), scaling factor \(\alpha=16\), and dropout \(0.05\), and inject adapters into the self-attention projection layers:
\(\{\texttt{q\_proj}, \texttt{k\_proj}, \texttt{v\_proj}, \texttt{o\_proj}\}\).
With the backbone frozen, only the LoRA parameters and the lightweight actor heads are optimized, yielding an efficient training setup.
\paragraph{Layer mixing for token representations.}
Instead of relying on a single transformer layer, we form token features using a learned convex combination of the last four hidden layers. Let \(\mathbf{h}_i^{(-4)},\dots,\mathbf{h}_i^{(-1)}\) denote the last four layer outputs. We learn a parameter vector \(\mathbf{a}\in\mathbb{R}^{4}\) and compute weights \(w_k=\mathrm{softmax}(\mathbf{a})_k\). The mixed token representation is:
\begin{equation}
\mathbf{x}_i = \sum_{k=1}^{4} w_k \, \mathbf{h}_i^{(-k)} .
\end{equation}
This aggregation often provides more robust saliency features than using only the final layer.
\paragraph{Policy and value heads.}
On top of \(\mathbf{x}_i\), we attach a lightweight linear head with a LayerNorm. The \emph{policy head} produces token-level logits \(z_i\) and converts them to Bernoulli importance probabilities via a temperature-controlled sigmoid:
\begin{equation}
p_i = \sigma\!\left(\frac{z_i}{\tau}\right),
\end{equation}
where \(\tau\) is the temperature and \(i \in \{1,\dots,L\}\) indexes token positions. These probabilities parameterize the actor's token-wise importance policy: \(p_i\) is the probability of marking token \(i\) as important.
\noindent
Overall, our actor combines a frozen pretrained LLM backbone with parameter-efficient LoRA adapters and a small policy head, yielding a compact and stable model for token-level emphasis learning.
\subsection{Data Preprocessing}

\paragraph{Amazon-Beauty.}
For each user, we parse the chronological interaction sequence (purchase history) as a list of item IDs
\(\mathcal{H}_u = [{item}_1, {item}_2, \ldots, {item}_T]\).
We formulate next-item recommendation by using the \emph{last purchased item} as the supervision label,
\(y_u = i_T\), and the remaining prefix as the model input history,
\(\mathcal{H}'_u = [{item}_1, \ldots, {item}_{T-1}]\).
Users with fewer than two interactions are discarded since they do not provide a valid (history, label) pair.

We build an item-item co-visitation (co-vis) graph from training histories using a sliding-window scheme.
For each user history \(\mathcal{H}_u\), we iterate over positions \(t\) and add weighted edges between item pairs that
co-occur within a window of radius \(v\):
\[
\forall\, j \in [t-v,\, t+v],\; j\neq t:\quad
\mathrm{co}[{item}_t][{item}_j] \mathrel{+}= 1.
\]
Aggregating over all users yields a directed adjacency map where each item \(a\) has a neighbor list
\(\mathcal{N}(a)=\{(b,\,\mathrm{co}[a][b])\}\) sorted by descending co-visitation count.

For each user history \(\mathcal{H}'_u\), we construct a candidate set \(\mathcal{C}_u\) by retrieving neighbors from the
co-vis graph, emphasizing recent interactions.
Let \(\mathcal{R}_u\) be the last \(M\) items of the history (we use \(M=10\)) in reverse chronological order.
We score each candidate item \(c\notin \mathcal{H}'_u\) by accumulating co-vis weights from recent items with an
exponential recency decay:
\[
s(c) = \sum_{r\in \mathcal{R}_u} \alpha^{\mathrm{rank}(r)} \cdot \mathrm{co}[r][c],
\]
where \(\alpha\in(0,1)\) controls the decay (we use \(\alpha=0.85\)) and \(\mathrm{rank}(r)=0\) denotes the most recent item.
We then take the top-\(K\) items by score as co-vis candidates. If fewer than \(K\) candidates are available, we fill the
remaining slots with the most popular items not appearing in the history nor already selected. By default, we select 40 candidates for both training and evaluation.

\textbf{Other datasets.} For other datasets, we use the default dataset loader from \footnote{https://huggingface.co/docs/datasets/en/index}{Hugging Face}.

\subsection{Hyperparameter Summary}
\label{app:hyperparams}

Table~\ref{tab:hyperparams} summarizes the key hyperparameters across tasks; we additionally fix the training batch size to 1 and the sliding-window radius to 3 for candidate-set construction.

\begin{table}[h]
\centering
\small
\begin{tabular}{llp{6cm}}
\toprule
\textbf{Symbol} & \textbf{Default} & \textbf{Description} \\
\midrule
$\tau$ & 1.0 & Temperature in Eq.~\eqref{eq:importance_prob} for scaling token-importance logits before the sigmoid \\
$\gamma$ & 0.25 (Beauty), 0.15 (others) & Highlight budget fraction; $k=\lfloor \gamma L \rfloor$ tokens selected after projection \\
$\delta$ & 10 & Gap-bridging threshold in span coalescence: spans separated by $\le \delta$ tokens are merged \\
$G$ & 4 & Number of sampled masks per instance in grouped policy-gradient training \\
$\Omega$ & All context tokens & Set of policy-controlled tokens (query tokens are excluded from highlighting) \\
$\lambda_{\text{len}}$ & 0.01 & Weight for target-length regularization (Eq.~\eqref{eq:len_reg}) \\
$\beta_{\text{ent}}$ & 1.0 & Entropy bonus coefficient (Eq.~\eqref{eq:ent_bonus}) \\
lr & $1\times10^{-4}$ & Adam learning rate \\
wd & $1\times10^{-2}$ & Weight decay \\
LoRA $r$ & 8 & LoRA rank for Actor backbone adaptation \\
LoRA $\alpha$ & 16 & LoRA scaling factor \\
\bottomrule
\end{tabular}
\caption{Consolidated hyperparameter table.}
\label{tab:hyperparams}
\end{table}

\subsection{Training Settings}
\paragraph{Amazon-Beauty.}
For the Amazon-Beauty recommendation task, the actor takes only the user’s historical reviews as input.
The candidate set (along with each item’s metadata, e.g., title/brand/category/price band and short phrases) is treated as part of the query used by the downstream solver, rather than being fed into the actor. In this way, the actor focuses on selecting salient preference signals from the user history, while the re-ranker conditions on both the emphasized history and the candidate descriptions.

\paragraph{Question answering.}
For QA tasks, we feed the actor a concatenation of the question and the evidence,
but we apply highlighting only to the evidence span. This design encourages the actor to identify the minimal supporting evidence for answering, while keeping the question intact as a conditioning prompt for the downstream QA model.

\begin{revblock}
\subsection{Training Protocol and Data Access}\label{app:protocol}

Table~\ref{tab:splits} lists the data source, training use, and evaluation set for each task. Actor training consumes a sequential prefix of each training source in a single pass. With batch size 1 and one RL update per training instance, Amazon-Beauty uses 3,000 unique training instances and 3,000 RL updates, while HotpotQA, SQuAD 2.0, and PubMedQA each use 10,000 unique training instances and 10,000 RL updates. Each update evaluates $G=4$ sampled masks, resulting in 12K optimization-time Solver queries for Amazon-Beauty and 40K for each QA task. Checkpoints are saved every 1K updates. For each dataset, checkpoint selection uses a fixed 500-example subset sampled with seed 42 from that dataset's training split; these examples are disjoint from both the instances used for Actor optimization and all final evaluation examples. In particular, none of the 7,405 HotpotQA distractor-validation examples used for final evaluation is used for checkpoint selection.

\begin{table}[h]
\color{\revcolor}
\centering
\footnotesize
\begin{tabular}{@{}>{\raggedright\arraybackslash}p{3.4cm}>{\raggedright\arraybackslash}p{4.7cm}>{\raggedright\arraybackslash}p{4.7cm}@{}}
\toprule
\textbf{Dataset and configuration} & \textbf{Training source and use} & \textbf{Evaluation source and use} \\
\midrule
Amazon-Beauty; final item for each user as the label; 40 co-visitation candidates & \texttt{train\_split}: 3,470,973 rows; first 3,000 used for Actor training & \texttt{test\_split}: 867,744 rows; users sampled (seed 42) until 1,000 have the ground truth among the 40 candidates; metrics over all 18,633 sampled user  \\
HotpotQA; \texttt{hotpot\_qa}, \texttt{distractor} configuration (Hugging Face) & 90,447 rows; first 10,000 processed sequentially in one pass & Full distractor validation split: 7,405 examples (official test labels are hidden) \\
SQuAD 2.0; \texttt{squad\_v2} (Hugging Face) & 130,319 rows; first 10,000 processed sequentially in one pass & 2,000 examples sampled from the validation split (seed 42) \\
PubMedQA; \texttt{pubmed\_qa}, \texttt{pqa\_artificial} configuration (Hugging Face) & 211,269 rows; first 10,000 processed sequentially in one pass & Fixed evaluation sample of 1,000 examples \\
\bottomrule
\end{tabular}
\caption{\rev{Data splits and their use for Actor training and evaluation.}}
\label{tab:splits}
\end{table}

Table~\ref{tab:budget} reports the optimization-time Solver queries of all methods on the four tasks, and Table~\ref{tab:data_access} summarizes their training-data access, supervision signals, and selection rules. All learned methods draw from the same per-task optimization pool. The prompt optimizers use larger Solver-query budgets and, depending on the method, observe gold demonstrations or meta-prompt feedback, whereas \HiLight receives only the scalar downstream reward, without in-context gold outputs or token-level evidence labels. MI and \textsc{Self-Mark} use no optimization data or optimization-time Solver queries, although \textsc{Self-Mark} adds one extra Solver query per example at inference time.

\begin{table}[h]
\color{\revcolor}
\centering
\small
\resizebox{\textwidth}{!}{
\begin{tabular}{@{}lcccccc@{}}
\toprule
\textbf{Method} & \textbf{Optimization or control} & \textbf{Instance-conditioned} & \textbf{Beauty} & \textbf{HotpotQA} & \textbf{SQuAD} & \textbf{PubMedQA} \\
\midrule
MI & Fixed manual instruction & No & 0 & 0 & 0 & 0 \\
PRL & Task-level learned prompt & No & 120K & 400K & 400K & 400K \\
BFRS & Prompt and demonstration search & No & 60K & 200K & 200K & 200K \\
DSPy/MIPROv2 & Prompt-program optimization & No & 24K & 80K & 80K & 80K \\
APE & Instruction induction & No & 60K & 200K & 200K & 200K \\
OPRO & Instruction optimization & No & 30K & 100K & 100K & 100K \\
\textsc{Self-Mark} & Zero-shot self-highlighting & Yes & 0 & 0 & 0 & 0 \\
\HiLight & Learned context-span emphasis & Yes & 12K & 40K & 40K & 40K \\
\bottomrule
\end{tabular}}
\caption{\rev{Optimization-time Solver queries for each method and task.}}
\label{tab:budget}
\end{table}

\begin{table}[h]
\color{\revcolor}
\centering
\scriptsize
\begin{tabular}{@{}l>{\raggedright\arraybackslash}p{2.3cm}>{\raggedright\arraybackslash}p{5.0cm}>{\raggedright\arraybackslash}p{3.4cm}@{}}
\toprule
\textbf{Method} & \textbf{Training-data access} & \textbf{Supervision signal} & \textbf{Selection basis} \\
\midrule
MI & none & none & fixed instruction \\
PRL & shared per-task subset & task-metric reward on candidate prompts; failure cases with gold answers shown in the meta-prompt & best instruction by subset score \\
BFRS & shared per-task subset & task metric; self-bootstrapped demonstrations (gold outputs of solved examples) & best-first search score on subset \\
DSPy/MIPROv2 & shared per-task subset & task metric via surrogate-guided evaluation; grounded demonstrations with gold outputs & best configuration by surrogate or evaluation score \\
APE & shared per-task subset & gold input and output demonstrations; candidate scoring on subset & highest-scoring instruction \\
OPRO & shared per-task subset & scores of prior candidates and gold demonstrations in the meta-prompt & top instruction by subset score \\
\textsc{Self-Mark} & none & none & not applicable (zero-shot) \\
\HiLight & shared per-task subset & outcome-level task reward only; no gold answers in context and no token-level evidence labels & checkpoint selected on a held-out subset \\
\bottomrule
\end{tabular}
\caption{\rev{Training-data access, supervision, and selection rules of each method.}}
\label{tab:data_access}
\end{table}
\end{revblock}

\subsection{Prompts}

In this section, we provide the prompts used in our experiments.

\begin{tcolorbox}[
  title=Solver LLM Re-ranking Prompt for Amazon-Beauty (Training),
  colback=white,
  colframe=black,
  boxrule=0.5pt,
  breakable
]
\begin{Verbatim}[fontsize=\small, breaklines=true]
You are a recommender re-ranker. Your goal is to score each candidate from 0 to 10 based on how likely it is to be the user’s next item. 

Return JSON only, for example:
<FINAL_JSON>
[{"id": 1,"score": 8.5}, {"id":2,"score":6.0}, {"id": 10, "score": 0.7}, {"id": 5, "score": 0.6}, ...]
</FINAL_JSON>

[USER_HISTORY_COMPACT]
<tagged_evidence>

[CANDIDATES]
cid=1 | title="<title>" | brand="<brand>" | cat="<cat>" | price_band="<price_band>" | rating=<avg_rating>(<rating_count>) | <phrases>
cid=2 | title="<title>" | brand="<brand>" | cat="<cat>" | price_band="<price_band>" | rating=<avg_rating>(<rating_count>) | <phrases>
...
Example Final JSON Output:

Rules:
1. Output ONLY the JSON array, nothing before or after.

2. Keep the format strictly valid JSON.
\end{Verbatim}
\end{tcolorbox}

\begin{tcolorbox}[
  title=Solver LLM Re-ranking Prompt for Amazon-Beauty (Testing),
  colback=white,
  colframe=black,
  boxrule=0.5pt,
  breakable
]
\begin{Verbatim}[fontsize=\small, breaklines=true]
Given a user's history summary and a list of candidate items, re-rank the candidates according to their likelihood of being the next item of interest to the user based on the user's historical interactions. Each entry should contain an "id" and a "score" reflecting the predicted relevance of the candidate to the user. Ensure that the JSON is well-formed and encapsulated within <FINAL_JSON> tags. Some parts of the history are wrapped in <start_important> ... <end_important>. Treat those spans as especially important signals about the user preferences.

Example Final JSON Output:

<FINAL_JSON>
[{"id": 1,"score": 8.5}, {"id":2,"score":6.0}, {"id": 10, "score": 0.7}, {"id": 5, "score": 0.6}, ...]
</FINAL_JSON>

Please adhere to the provided structure and ensure the JSON output is strictly formatted as shown.

[USER_HISTORY_SUMMARY]
<user_summary>

[CANDIDATES]
id=1 | title="<title>" | brand="<brand>" | cat="<cat>" | price_band="<price_band>" | rating=<avg_rating>(<rating_count>) | <phrases>
id=2 | title="<title>" | brand="<brand>" | cat="<cat>" | price_band="<price_band>" | rating=<avg_rating>(<rating_count>) | <phrases>

Final JSON Output:
\end{Verbatim}
\end{tcolorbox}

\begin{tcolorbox}[
  title=QA tasks Training-Time Prompt,
  colback=white,
  colframe=black,
  boxrule=0.5pt,
  breakable
]
\begin{Verbatim}[fontsize=\small, breaklines=true]
You are a helpful, precise QA assistant.
Follow the format EXACTLY:
You MUST output ONLY the short answer phrase inside <answer>...</answer>.
No explanation, no extra words.
Some parts of the EVIDENCE are wrapped in <start_important> ... <end_important>.

QUESTION:
<question>

EVIDENCE:
<evidence>

OUTPUT:
\end{Verbatim}
\end{tcolorbox}

\begin{tcolorbox}[
  title=QA tasks Testing-Time Prompt,
  colback=white,
  colframe=black,
  boxrule=0.5pt,
  breakable
]
\begin{Verbatim}[fontsize=\small, breaklines=true]
You are a helpful, precise QA assistant.
Follow the format EXACTLY:
You MUST output ONLY the short answer phrase inside <answer>...</answer>.
No explanation, no extra words.
Some parts of the EVIDENCE are wrapped in <start_important> ... <end_important>.

QUESTION:
<question>

EVIDENCE:
<evidence>

OUTPUT:
\end{Verbatim}
\end{tcolorbox}

\begin{tcolorbox}[
  title=OPRO Meta-Prompt for Amazon-Beauty,
  colback=white,
  colframe=black,
  boxrule=0.5pt,
  breakable
]
\begin{Verbatim}[fontsize=\small, breaklines=true]
You are an expert prompt engineer optimizing an instruction for a recommender reranking task.
The instruction will be used in a 0-shot prompt that asks the model to output ONLY:
[{"id": 1, "score": 7.5}, {"id": 2, "score": 3.0}, ...]

The model sees:
- A user history summary (text describing past interactions)
- A list of candidate items with metadata (title, brand, category, price, rating, review phrases)

Your job: propose {n} new instruction variants that might improve HR@10 and NDCG@10.

[TOP INSTRUCTIONS SO FAR]
1. "{instruction_1}" → HR@10={hr1:.3f}, NDCG@10={ndcg1:.3f}
2. "{instruction_2}" → HR@10={hr2:.3f}, NDCG@10={ndcg2:.3f}
...

Generate {n} new diverse instructions. Output each on its own line, no numbering, no quotes.
\end{Verbatim}
\end{tcolorbox}

\begin{tcolorbox}[
  title=OPRO Meta-Prompt for QA tasks,
  colback=white,
  colframe=black,
  boxrule=0.5pt,
  breakable
]
\begin{Verbatim}[fontsize=\small, breaklines=true]
You are an expert prompt engineer.
Goal: write an INSTRUCTION for question answering.
The instruction will be placed BEFORE QUESTION and EVIDENCE.

Hard constraints:

Answer using ONLY the provided EVIDENCE.
Be concise.
Output ONLY: <FINAL_ANSWER>answer</FINAL_ANSWER> (no extra text, no markdown, no headings).

Here are example INPUT/OUTPUT pairs (demos). Your instruction should make the model produce outputs like these:

===== DEMO 1 INPUT =====
QUESTION:
<demo_1_question>

EVIDENCE:
<demo_1_evidence>

===== DEMO 1 OUTPUT =====
<FINAL_ANSWER>
<demo_1_answer>
</FINAL_ANSWER>
...

Best instructions so far (higher F1/EM is better):

--- BEST 1 ---
F1=<f1> EM=<em> empty=<empty> n=<n>
<best_instruction_1>

--- BEST 2 ---
...

Now generate <n_new> NEW, diverse instruction candidates.
Return ONLY valid JSON (no markdown) in this schema:
{"instructions": ["...", "...", ...]}
Do not include any other keys. Do not include explanations.
\end{Verbatim}
\end{tcolorbox}

\begin{tcolorbox}[
  title=PRL Meta-Prompt for Amazon-Beauty,
  colback=white,
  colframe=black,
  boxrule=0.5pt,
  breakable
]
\begin{Verbatim}[fontsize=\small, breaklines=true]
You are an expert prompt engineer.

Task: Optimize an instruction for a recommender system that re-ranks candidate items.

The instruction will be inserted into a prompt template that includes:
- User history summary
- Candidate items with metadata

The model must output a JSON array: [{"id": int, "score": float}, ...]

[CURRENT TOP INSTRUCTIONS]
{ranked_instructions_with_scores}

[FEEDBACK]
{optional_feedback_from_failed_cases}

Generate {n} new instruction variants that could improve Hit Rate and NDCG metrics.
Be creative but maintain the core task requirements.
\end{Verbatim}
\end{tcolorbox}

\begin{tcolorbox}[
  title=PRL Meta-Prompt for QA tasks,
  colback=white,
  colframe=black,
  boxrule=0.5pt,
  breakable
]
\begin{Verbatim}[fontsize=\small, breaklines=true]
You are an expert prompt optimizer.
Task: improve an INSTRUCTION for question answering.
The instruction will be placed BEFORE QUESTION and EVIDENCE.

Hard constraints for ANY instruction you propose:

The model must answer based on ONLY the provided EVIDENCE.

Output ONLY: <FINAL_ANSWER>answer</FINAL_ANSWER> (no extra text, no markdown, no headings).

Be concise (often 1–8 words).
Current instruction:
----- CURRENT_INSTRUCTION -----
<current_instr>
----- END_CURRENT_INSTRUCTION -----

Current score on training subset: metric(F1+0.1*EM)=<metric> F1=<f1> EM=<em> empty=<empty> n=<n>

Top instructions so far (for context):
[TOP 1] metric=<metric> F1=<f1> EM=<em>
<top_instruction_1>
[TOP 2] ...

Here are some failure cases of the CURRENT instruction. Improve the instruction to fix these:
===== FAILURE 1 =====
QUESTION:
<failure_1_question>

EVIDENCE:
<failure_1_evidence>

GOLD:
<failure_1_gold>

PRED:
<failure_1_pred>

F1=<f1> EM=<em>
===== FAILURE 2 =====
...
Now propose <n_new> improved, diverse instruction candidates.
Each candidate should be a SMALL EDIT of the current instruction (not a completely different format), but you may add 1–3 extra rules that improve answer extraction and concision.

Return ONLY valid JSON (no markdown) in this schema:
{"instructions": ["...", "...", ...]}
No other keys. No explanations.
\end{Verbatim}
\end{tcolorbox}

\begin{tcolorbox}[
  title=DSPy Base-Instruction for Amazon-Beauty,
  colback=white,
  colframe=black,
  boxrule=0.5pt,
  breakable
]
\begin{Verbatim}[fontsize=\small, breaklines=true]
You are a recommendation re-ranker.
You will be given two blocks:
[USER_HISTORY_SUMMARY] ...
[CANDIDATES] lines with cid=... and metadata

Score EACH candidate by likelihood of being the user's NEXT interaction.
Use a 0–10 scale (higher = more likely) with meaningful spread (avoid ties).

STRICT output format (no extra text):
<FINAL_JSON>
[{"id": cid_int, "score": number}, ...]
</FINAL_JSON>

Rules:

Include EVERY candidate exactly once; id must match the provided cid.
Use user history signals (brand/category/phrases) to guide scores.
\end{Verbatim}
\end{tcolorbox}

\begin{tcolorbox}[
  title=DSPy Base-Instruction for QA tasks,
  colback=white,
  colframe=black,
  boxrule=0.5pt,
  breakable
]
\begin{Verbatim}[fontsize=\small, breaklines=true]
Question answering task.
You will be given a QUESTION and EVIDENCE context.
Answer the question using ONLY the evidence. Be concise.

STRICT output format (no extra text):
<FINAL_ANSWER>
your answer here
</FINAL_ANSWER>
\end{Verbatim}
\end{tcolorbox}

\begin{tcolorbox}[
  title=APE-style Instruction Induction for Amazon-Beauty,
  colback=white,
  colframe=black,
  boxrule=0.5pt,
  breakable
]
\begin{Verbatim}[fontsize=\small, breaklines=true]
Goal: write an INSTRUCTION for a recommender re-ranking task.
This instruction will be placed BEFORE the blocks [USER_HISTORY_SUMMARY] and [CANDIDATES].

The model must output ONLY this exact wrapper and JSON:
<FINAL_JSON>
[{"id": cid_int, "score": number}, ...]
</FINAL_JSON>

Hard constraints for your instruction:

Enforce strict output formatting: ONLY <FINAL_JSON>...</FINAL_JSON> with a JSON array.

Include EVERY candidate exactly once; "id" is the provided cid integer; "score" is a number.

Encourage meaningful score spread (avoid ties) and use user history signals.

Keep the instruction concise (<= 120 words).

Here are example INPUT/OUTPUT pairs (demos). The instruction you write should make the model produce outputs like these:

===== DEMO 1 INPUT =====
<demo_1_input>
===== DEMO 1 OUTPUT =====
<demo_1_output>

===== DEMO 2 INPUT =====
<demo_2_input>
===== DEMO 2 OUTPUT =====
<demo_2_output>

...

Here are the best instructions found so far and their evaluation metrics (higher NDCG/HR is better):

--- BEST 1 ---
NDCG=<ndcg> HR=<hr> empty_parse=<empty_parse> n=<n>
<best_instruction_1>

--- BEST 2 ---
NDCG=<ndcg> HR=<hr> empty_parse=<empty_parse> n=<n>
<best_instruction_2>

...

Now generate <n_new> NEW, diverse instruction candidates.
Return ONLY valid JSON (no markdown) in this schema:
{"instructions": ["...", "...", ...]}
Do not include any other keys. Do not include explanations.
\end{Verbatim}
\end{tcolorbox}

\begin{tcolorbox}[
  title=APE-style Instruction Induction for QA tasks,
  colback=white,
  colframe=black,
  boxrule=0.5pt,
  breakable
]
\begin{Verbatim}[fontsize=\small, breaklines=true]
You are an expert prompt engineer (APE).
Goal: write an INSTRUCTION for question answering.
The instruction will be placed BEFORE QUESTION and EVIDENCE.

Hard constraints:

Answer using ONLY the provided EVIDENCE.

Output ONLY: <FINAL_ANSWER>answer</FINAL_ANSWER> (no extra text, no markdown).

Be concise; prefer a short span/phrase or "yes"/"no" when appropriate.

Examples (demos):

===== DEMO 1 INPUT =====
QUESTION:
<demo_1_question>

EVIDENCE:
<demo_1_evidence>

===== DEMO 1 OUTPUT =====
<FINAL_ANSWER>
<demo_1_answer>
</FINAL_ANSWER>

===== DEMO 2 INPUT =====
QUESTION:
<demo_2_question>

EVIDENCE:
<demo_2_evidence>

===== DEMO 2 OUTPUT =====
<FINAL_ANSWER>
<demo_2_answer>
</FINAL_ANSWER>

...

Best instructions so far (higher metric = F1 + 0.1*EM is better):

--- BEST 1 ---
metric=<metric> F1=<f1> EM=<em> empty=<empty> n=<n>
<best_instruction_1>

--- BEST 2 ---
metric=<metric> F1=<f1> EM=<em> empty=<empty> n=<n>
<best_instruction_2>

...

Now generate <n_new> NEW, diverse instruction candidates.
Return ONLY valid JSON (no markdown) in this schema:
{"instructions": ["...", "...", ...]}
No other keys. No explanations.
\end{Verbatim}
\end{tcolorbox}

\begin{tcolorbox}[
  title=BFRS Prompt for Amazon-Beauty,
  colback=white,
  colframe=black,
  boxrule=0.5pt,
  breakable
]
\begin{Verbatim}[fontsize=\small, breaklines=true]
You are a recommender re-ranker.
Goal: score each candidate 0–10 for next-item likelihood for this user.

Output format rules:

You MUST output ONLY the final JSON array inside <FINAL_JSON>...</FINAL_JSON>.

<FINAL_JSON> must be a single valid JSON array of objects with fields "id" and "score".

No extra text outside <FINAL_JSON>.

The "id" field must be an integer matching id below.

The "score" field must be a number (int or float).

Here are some solved examples:

EXAMPLE 1

[USER_HISTORY_SUMMARY]
<example_1_user_summary>

[CANDIDATES]
id=1 | title="<...>" | brand="<...>" | cat="<...>" | price_band="<...>" | rating=<...>(<...>) | <phrases>
id=2 | title="<...>" | brand="<...>" | cat="<...>" | price_band="<...>" | rating=<...>(<...>) | <phrases>
...

<FINAL_JSON>
<example_1_gold_json>
</FINAL_JSON>

EXAMPLE 2

[USER_HISTORY_SUMMARY]
<example_2_user_summary>

[CANDIDATES]
cid=1 | title="<...>" | brand="<...>" | cat="<...>" | price_band="<...>" | rating=<...>(<...>) | <phrases>
cid=2 | title="<...>" | brand="<...>" | cat="<...>" | price_band="<...>" | rating=<...>(<...>) | <phrases>

<FINAL_JSON>
<example_2_gold_json>
</FINAL_JSON>

Now solve the next case:

[USER_HISTORY_SUMMARY]
<user_summary>

[CANDIDATES]
cid=1 | title="<title>" | brand="<brand>" | cat="<cat>" | price_band="<price_band>" | rating=<avg_rating>(<rating_count>) | <phrases>
cid=2 | title="<title>" | brand="<brand>" | cat="<cat>" | price_band="<price_band>" | rating=<avg_rating>(<rating_count>) | <phrases>
...
Output:
\end{Verbatim}
\end{tcolorbox}

\begin{tcolorbox}[
  title=BFRS Prompt for QA tasks,
  colback=white,
  colframe=black,
  boxrule=0.5pt,
  breakable
]
\begin{Verbatim}[fontsize=\small, breaklines=true]
You are an expert prompt optimizer (BFRS).
Task: improve an INSTRUCTION for question answering.
The instruction will be placed BEFORE QUESTION and EVIDENCE.

Hard constraints for ANY instruction you propose:

The model must answer using ONLY the provided EVIDENCE.
Output ONLY: <FINAL_ANSWER>answer</FINAL_ANSWER> (no extra text, no markdown, no headings).
Be concise (often 1–8 words); prefer a short span/phrase or "yes"/"no" when appropriate.

Current best instruction:
----- CURRENT_INSTRUCTION -----
<current_instr>
----- END_CURRENT_INSTRUCTION -----

Current score on training subset: metric(F1+0.1*EM)=<metric> F1=<f1> EM=<em> empty=<empty> n=<n>

Top instructions so far (for context):
[TOP 1] metric=<metric> F1=<f1> EM=<em>
<top_instruction_1>

[TOP 2] metric=<metric> F1=<f1> EM=<em>
<top_instruction_2>

...

Here are some failure cases of the CURRENT instruction. Improve the instruction to fix these:

===== FAILURE 1 =====
QUESTION:
<failure_1_question>

EVIDENCE:
<failure_1_evidence>

GOLD:
<failure_1_gold>

PRED:
<failure_1_pred>

F1=<f1> EM=<em>

===== FAILURE 2 =====
...

BFRS generation rule:

Start from the CURRENT instruction and TOP instructions as “parents”.
Create each new candidate by a SMALL EDIT (do not change the output tag format).
You may add 1–3 extra rules that improve answer extraction (shortest span, exact match, yes/no) and reduce empty/wrong-format outputs.
Ensure diversity across candidates (different phrasing/rules), while staying concise.

Now propose <n_new> improved, diverse instruction candidates.
Return ONLY valid JSON (no markdown) in this schema:
{"instructions": ["...", "...", ...]}
No other keys. No explanations.
\end{Verbatim}
\end{tcolorbox}

\end{document}